\documentclass{article}
\usepackage{iclr2024_conference,times}

\usepackage{array}
\usepackage[T1]{fontenc}
\usepackage[utf8]{inputenc}
\usepackage{graphicx}
\usepackage{booktabs}
\usepackage{amsmath}
\usepackage{amssymb}
\usepackage{amsfonts}
\usepackage{multirow}
\usepackage{verbatim}
\usepackage{caption}
\usepackage{longtable}
\usepackage{supertabular}
\usepackage{float}
\usepackage{mathtools}
\usepackage{latexsym}

\usepackage{enumitem}
\usepackage{tablefootnote}
\usepackage{xcolor}
\usepackage{xspace}
\usepackage{textcomp}
\usepackage{makecell}
\usepackage{lscape} 
\usepackage{siunitx}

\usepackage{scrextend}

\usepackage{microtype}
\definecolor{mydarkblue}{rgb}{0,0.08,0.45}
\usepackage[colorlinks,citecolor=mydarkblue,urlcolor=mydarkblue,linkcolor=mydarkblue]{hyperref}
\usepackage{url}            %
\usepackage{nicefrac}       %
\usepackage{changepage}
\usepackage{xargs}          %
\usepackage{wrapfig,lipsum,booktabs}

\usepackage{subcaption}
\usepackage{endnotes}

\usepackage{pgfplots}
\usetikzlibrary{pgfplots.groupplots}
\pgfplotsset{compat=1.3}
\usepackage{tikz}
\usetikzlibrary{patterns}
\usepackage{sidecap}

\usepackage[most]{tcolorbox}

\usepackage[capitalize,noabbrev]{cleveref}
\crefname{section}{Section}{Sections}
\Crefname{section}{Section}{Sections}
\crefname{table}{Table}{Tables}
\crefname{figure}{Figure}{Figures}
\crefname{algorithm}{Algorithm}{Algorithms}
\crefname{equation}{eq.}{eqs.}
\crefname{appendix}{Appendix}{Appendices}
\crefformat{section}{Section #2#1#3}
\usepackage{multicol}

\usepackage{minitoc}

\title{From Sparse to Soft Mixtures of Experts}

\author{%
Joan Puigcerver\thanks{Equal contribution. The order was decided by a coin toss.}\\
Google DeepMind\\
\And
Carlos Riquelme\footnotemark[1] \\
Google DeepMind\\
\And
Basil Mustafa \\
Google DeepMind\\
\And
Neil Houlsby \\
Google DeepMind\\
}

\def\R{\mathbb{R}}
\def\mC{\mathbf{C}}
\def\mD{\mathbf{D}}
\def\mX{\mathbf{X}}
\def\mY{\mathbf{Y}}
\def\mPhi{\mathbf{\Phi}}

\definecolor{codegreen}{rgb}{0,0.6,0}
\definecolor{codegray}{rgb}{0.5,0.5,0.5}
\definecolor{codepurple}{rgb}{0.58,0,0.82}
\definecolor{backcolour}{rgb}{0.95,0.95,0.92}

\lstdefinestyle{mystyle}{
  backgroundcolor=\color{backcolour}, commentstyle=\color{codegreen},
  keywordstyle=\color{magenta},
  numberstyle=\tiny\color{codegray},
  stringstyle=\color{codepurple},
  basicstyle=\ttfamily\footnotesize,
  breakatwhitespace=false,         
  breaklines=true,                 
  captionpos=b,                    
  keepspaces=true,                 
  numbers=left,                    
  numbersep=5pt,                  
  showspaces=false,                
  showstringspaces=false,
  showtabs=false,                  
  tabsize=2
}

\crefname{listing}{algorithm}{algorithms}
\Crefname{listing}{Algorithm}{Algorithms}

\newcommand{\name}{\mbox{Soft MoE}\xspace}
\newcommand{\names}{\mbox{Soft MoEs}\xspace}

\iclrfinalcopy
\begin{document}
\setlength{\abovedisplayskip}{4pt}
\setlength{\belowdisplayskip}{4pt}
\setlength{\abovedisplayshortskip}{0pt}
\setlength{\belowdisplayshortskip}{0pt}

\doparttoc %
\faketableofcontents %

\maketitle

\begin{abstract}
Sparse mixture of expert architectures (MoEs) scale model capacity without significant increases in training or inference costs.
Despite their success, MoEs suffer from a number of issues: training instability, token dropping, inability to scale the number of experts, or ineffective finetuning.
In this work, we propose \name, a \emph{fully-differentiable} sparse Transformer that addresses these challenges, while maintaining the benefits of MoEs.
\name performs an implicit soft assignment by passing different weighted combinations of all input tokens to each expert.
As in other MoEs, experts in \name only process a subset of the (combined) tokens, enabling larger model capacity (and performance) at lower inference cost.
In the context of visual recognition, \name greatly outperforms dense Transformers (ViTs) and popular MoEs (Tokens Choice and Experts Choice).
Furthermore, \name scales well: \name Huge/14 with 128 experts in 16 MoE layers has over $40\times$ more parameters than ViT Huge/14, with only 2\% increased inference time, and substantially better quality.
\end{abstract}

\section{Introduction}

Larger Transformers improve performance at increased computational cost.
Recent studies suggest that model size and training data must be scaled together to optimally use any given training compute budget~\citep{kaplan2020scaling,hoffmann2022training,zhai2022scaling}.
A promising alternative that allows to scale models in size without paying their full computational cost is sparse mixtures of experts (MoEs). 
Recently, a number of successful approaches have proposed ways to sparsely activate token paths across the network in language~\citep{lepikhin2020gshard, fedus2022switch}, vision~\citep{riquelme2021scaling}, and multimodal models~\citep{mustafa2022multimodal}.

Sparse MoE Transformers involve a discrete optimization problem to decide which modules should be applied to each token.
These modules are commonly referred to as \emph{experts} and are usually MLPs.
Many techniques have been devised to find good token-to-expert matches: linear programs \citep{lewis2021base}, reinforcement learning \citep{bengio2015conditional}, deterministic fixed rules \citep{roller2021hash}, optimal transport \citep{liu2022sparsity}, greedy top-$k$ experts per token \citep{shazeer2017outrageously}, or greedy top-$k$ tokens per expert \citep{zhou2022mixture}.
Often, heuristic auxiliary losses are required to balance utilization of experts and minimize unassigned tokens.
These challenges can be greater in out-of-distribution settings: small inference batch sizes, novel inputs, or in transfer learning.

We introduce \name, that overcomes many of these challenges.
Rather than employing a sparse and discrete router that tries to find a good \emph{hard} assignment between tokens and experts, \names instead perform a \emph{soft} assignment by mixing tokens.
In particular, we compute several weighted averages of all tokens---with weights depending on both tokens and experts---and then we process each weighted average by its corresponding expert.

\name L/16 outperforms ViT H/14 on upstream, few-shot and finetuning while requiring almost half the training time, and being \textbf{2$\times$ faster at inference}.
Moreover, \name B/16 matches ViT H/14 on few-shot and finetuning and outperforms it on upstream metrics after a comparable amount of training. Remarkably, \name B/16 is \textbf{5.7$\times$ faster at inference} despite having 5.5$\times$ the number of parameters of ViT H/14 (see \cref{tab:long_runs} and \cref{fig:long_runs} for details).
\Cref{sec:contrastive_experiments} demonstrates \name's potential to extend to other tasks: we train a contrastive model text tower against the frozen vision tower, showing that representations learned via soft routing preserve their benefits for image-text alignment.

\section{Soft Mixture of Experts}
\label{sec:softmoe}
\subsection{Algorithm description}
\label{sec:softmoe_description}

\begin{figure}[tb]
\centering
\includegraphics[width=\textwidth]{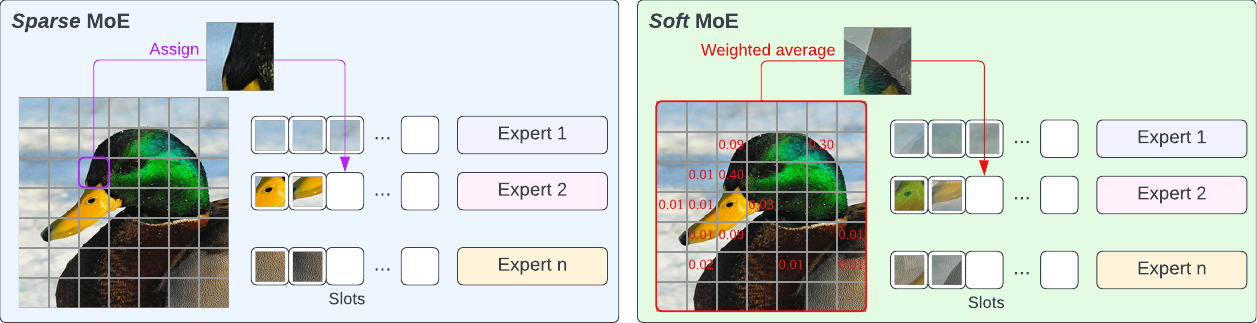}
\caption{%
\textbf{Sparse and \name layers.} While the router in Sparse MoE layers (left) learns to \emph{assign} individual input tokens to each of the available slots, in \name layers (right) each slot is the result of a (different) \emph{weighted average} of all the input tokens. Learning to make discrete assignments introduces several optimization and implementation issues that \name sidesteps.
\Cref{app:model_inspection} visualizes learned distributions of soft-assignments by \name.
\label{fig:sparse_vs_soft_diagram}}
\end{figure}

\begin{SCfigure}
\centering
\includegraphics[width=0.7\textwidth]{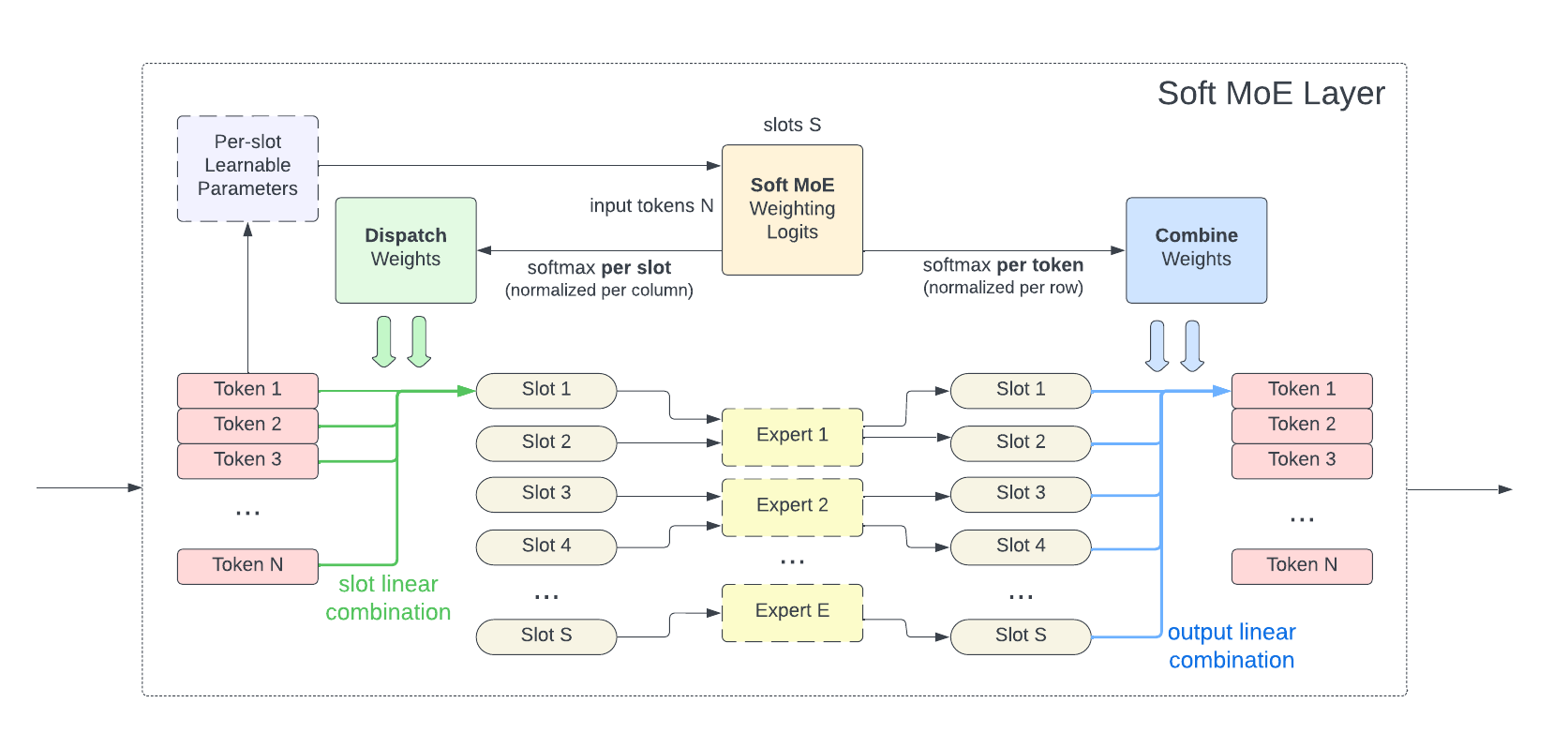}
\caption{%
\textbf{\name routing details.}
\name computes scores or logits for every pair of input token and slot.
From this it computes a slots$\times$tokens matrix of logits, that are normalized appropriately to compute both the dispatch and combine weights.
The slots themselves are allocated to experts round-robin.
\label{fig:soft_moe_diagram}}
\end{SCfigure}

The \name routing algorithm is depicted in \cref{fig:soft_moe_diagram}.
We denote the inputs tokens for one sequence by $\mX \in \R^{m \times d}$, where $m$ is the number of tokens and $d$ is their dimension.
Each MoE layer uses a set of $n$ expert functions\footnote{%
In practice, all experts apply the same function with different parameters, usually an MLP.}
applied on individual tokens, namely $\{f_i: \R^d \rightarrow \R^d \}_{1:n}$. 
Each expert processes $p$ \emph{slots}, and each slot has a corresponding $d$-dimensional vector of parameters, $\mPhi \in \R^{d \times (n \cdot p)}$. 

In particular, the input slots $\tilde{\mX} \in \R^{(n \cdot p) \times d}$ are the result of convex combinations of all the $m$ input tokens, $\mX$:
\begin{equation}
\label{eq:dispatch_weights_def}
\begin{gathered}
\mD_{ij} = \frac{\exp((\mX \mPhi)_{ij})}{\sum_{i'=1}^m \exp((\mX \mPhi)_{i'j})},\qquad
\tilde{\mX} = \mD^\top \mX.
\end{gathered}
\end{equation}

Notice that $\mD$, which we call the \emph{dispatch} weights, is simply the result of applying a softmax over the \emph{columns} of $\mX \mPhi$.
Then, as mentioned above, the corresponding expert function is applied on each slot (i.e. on rows of $\tilde{\mX}$) to obtain the output slots:
$\tilde{\mY}_i = f_{\left\lfloor{i / p}\right\rfloor}(\tilde{\mX}_i)$.

Finally, the output tokens $\mY$ are computed as a convex combination of all ($n \cdot p$) output slots, $\tilde{\mY}$, whose weights are computed similarly as before:
\begin{equation}
\label{eq:combine_weights_def}
\begin{gathered}
\mC_{ij} = \frac{\exp((\mX \mPhi)_{ij})}{\sum_{j'=1}^{n \cdot p} \exp((\mX \mPhi)_{ij'})},\qquad
\mY = \mC \tilde{\mY}.
\end{gathered}
\end{equation}
We refer to $\mC$ as the \emph{combine} weights, and it is the result of applying a softmax over the \emph{rows} of $\mX \mPhi$.

Following the usual design for Sparse MoEs, we replace a subset of the Transformer's MLP blocks with \name blocks. We typically replace the second half of MLP blocks.
The total number of slots is a key hyperparameter of \name layers because the time complexity depends on the number of slots rather than on the number of experts.
One can set the number of slots equal to the input sequence length to match the FLOPs of the equivalent dense Transformer.

\subsection{Properties of \name and connections with Sparse MoEs}
\label{sec:soft_moe_vs_sparse}

\paragraph{Fully differentiable}
Sparse MoE algorithms involve an assignment problem between tokens and experts, which is subject to capacity and load-balancing constraints.
Different algorithms approximate the solution in different ways: for example, the top-$k$ or ``Token Choice'' router \citep{shazeer2017outrageously,lepikhin2020gshard,riquelme2021scaling} selects the top-$k$-scored experts for each token, while there are slots available in such expert (i.e.\ the expert has not filled its \emph{capacity}). The ``Expert Choice'' router \citep{zhou2022mixture} selects the top-\emph{capacity}-scored tokens for each expert. Other works suggest more advanced (and often costly) algorithms to compute the assignments, such as approaches based on Linear Programming algorithms \citep{lewis2021base}, Optimal Transport \citep{liu2022sparsity,clark2022unified} or Reinforcement Learning \citep{clark2022unified}.
Nevertheless virtually all of these approaches are discrete in nature, and thus non-differentiable.
In contrast, all operations in \name layers are continuous and fully differentiable. 
We can interpret the weighted averages with softmax scores as \emph{soft} assignments, rather than the \emph{hard} assignments used in Sparse MoE.

\begin{lstlisting}[language=Python, caption={Simple JAX \citep{bradbury2018jax} implementation of a \name layer. Full code is available at \url{https://github.com/google-research/vmoe}.}, label=alg:soft_moe_python, escapechar=|]
def soft_moe_layer(X, Phi, experts):
  # Compute the dispatch and combine weights.
  logits = jnp.einsum('md,dnp->mnp', X, Phi)|\label{line:logits}|
  D = jax.nn.softmax(logits, axis=(0,))
  C = jax.nn.softmax(logits, axis=(1, 2))
  # The input slots are a weighted average of all the input tokens,
  # given by the dispatch weights.
  Xs = jnp.einsum('md,mnp->npd', X, D)
  # Apply the corresponding expert function to each input slot.
  Ys = jnp.stack([
    f_i(Xs[i, :, :]) for i, f_i in enumerate(experts)],
    axis=0)
  # The output tokens are a weighted average of all the output slots,
  # given by the combine weights.
  Y = jnp.einsum('npd,mnp->md', Ys, C)
  return Y
\end{lstlisting}

\paragraph{No token dropping and expert unbalance}
The classical routing mechanisms tend to suffer from issues such as ``token dropping'' (i.e.\ some tokens are not assigned to any expert),
or ``expert unbalance'' (i.e.\ some experts receive far more tokens than others).
Unfortunately, performance can be severely impacted as a consequence.
For instance, the popular top-$k$ or ``Token Choice'' router \citep{shazeer2017outrageously} suffers from both,
while the ``Expert Choice'' router \citep{zhou2022mixture} only suffers from the former (see \cref{app:dropping} for some experiments regarding dropping).
\names are immune to token dropping and expert unbalance since every slot is filled with a weighted average of all tokens.

\paragraph{Fast}
The total number of slots determines the cost of a \name layer.
Every input applies such number of MLPs.
The total number of \emph{experts} is irrelevant in this calculation: few experts with many slots per expert or many experts with few slots per expert will have matching costs if the total number of slots is identical.
The only constraint we must meet is that the number of slots has to be greater or equal to the number of experts (as each expert must process at least one slot).
The main advantage of \name is completely avoiding sort or top-$k$ operations which are slow and typically not well suited for hardware accelerators.
As a result, \name is significantly \emph{faster} than most sparse MoEs (\cref{fig:heatmap}).
See \cref{sec:implementation} for time complexity details.

\paragraph{Features of both sparse and dense}
The \emph{sparsity} in Sparse MoEs comes from the fact that expert parameters are only applied to a subset of the input tokens.
However, \names are not technically sparse, since every slot is a weighted average of all the input tokens.
Every input token \emph{fractionally} activates all the model parameters.
Likewise, all output tokens are fractionally dependent on all slots (and experts). 
Finally, notice also that \names are not Dense MoEs, where every expert processes all input tokens, since every expert only processes a subset of the slots.

\paragraph{Per-sequence determinism}
Under capacity constraints, all Sparse MoE approaches route tokens in \emph{groups} of a fixed size and enforce (or encourage) balance within the group.
When groups contain tokens from different sequences or inputs, these tokens \emph{compete} for available spots in expert buffers.
Therefore, the model is no longer deterministic at the sequence-level, but only at the batch-level.
Models using larger groups tend to provide more freedom to the routing algorithm and usually perform better, but their computational cost is also higher.

\subsection{Implementation}
\label{sec:implementation}

\paragraph{Time complexity} Assume the per-token cost of a single expert function is $O(k)$. The time complexity of a \name layer is then $O(mnpd + npk)$. %
By choosing $p = O({m}/{n})$ slots per expert, i.e. the number of tokens over the number of experts, the cost reduces to $O(m^2 d + m k)$. 
Given that each expert function has its own set of parameters, increasing the number of experts $n$ and scaling $p$ accordingly, allows us to increase the total number of parameters without any impact on the time complexity.
Moreover, when the cost of applying an expert is large, the $mk$ term dominates over $m^2 d$, and the overall cost of a \name layer becomes comparable to that of applying a single expert on all the input tokens.
Finally, even when $m^2 d$ is not dominated, this is the same as the (single-headed) self-attention cost, thus it does not become a bottleneck in Transformer models. This can be seen in the bottom plot of \cref{fig:heatmap} where the throughput of \name barely changes when the number of experts increases from 8 to 4\,096 experts, while Sparse MoEs take a significant hit.

\paragraph{Normalization} In Transformers, MoE layers are typically used to replace the feedforward layer in each encoder block. Thus, when using pre-normalization as most modern Transformer architectures \citep{domhan2018much,xiong2020layer,riquelme2021scaling,fedus2022switch}, the inputs to the MoE layer are ``layer normalized''. This causes stability issues when scaling the model dimension $d$, since the softmax approaches a one-hot vector as $d \rightarrow \infty$ (see \cref{sec:layernorm_problems}). Thus, in \cref{line:logits} of \cref{alg:soft_moe_python} we replace \lstinline{X} and \lstinline{Phi} with \lstinline{l2_normalize(X, axis=1)} and \lstinline{scale * l2_normalize(Phi, axis=0)}, respectively; where \lstinline{scale} is a trainable scalar, and \lstinline{l2_normalize} normalizes the corresponding axis to have unit (L2) norm, as \Cref{alg:l2_normalization} shows.

\begin{lstlisting}[language=Python, caption={JAX implementation of the L2 normalization used in \name layers.}, label=alg:l2_normalization]
def l2_normalize(x, axis, eps=1e-6):
  norm = jnp.sqrt(jnp.square(x).sum(axis=axis, keepdims=True))
  return x * jnp.reciprocal(norm + eps)
\end{lstlisting}

For relatively small values of $d$, the normalization has little impact on the model's quality. 
However, with the proposed normalization in the \name layer, we can make the model dimension bigger and/or increase the learning rate (see \cref{sec:layernorm_problems}).

\paragraph{Distributed model} When the number of experts increases significantly, it is not possible to fit the entire model in memory on a single device, especially during training or when using MoEs on top of large model backbones. In these cases, we employ the standard techniques to distribute the model across many devices, as in \citep{lepikhin2020gshard,riquelme2021scaling,fedus2022switch} and other works training large MoE models. Distributing the model typically adds an overhead in the cost of the model, which is not captured by the time complexity analysis based on FLOPs that we derived above. In order to account for this difference, in all of our experiments we measure not only the FLOPs, but also the wall-clock time in TPUv3-chip-hours.

\section{Image Classification Experiments}
\label{sec:classification_experiments}

\textbf{Training Pareto frontiers}. In \cref{sec:pareto_plots} we compare dense ViT models at the Small, Base, Large and Huge sizes with their dense and sparse counterparts based on both Tokens Choice and Experts Choice sparse routing.
We study performance at different training budgets and show that \name dominates other models in terms of performance at a given training cost or time.

\textbf{Inference-time optimized models}. In \cref{sec:long_runs}, we present longer training runs (``overtraining'').
Relative to ViT, \name brings large improvements in terms of inference speed for a fixed performance level (smaller models: S, B) and absolute performance (larger models: L, H).

\textbf{Model ablations}. In \cref{sec:analysis,sec:abaltions} we investigate the effect of changing slot and expert counts, and perform ablations on the \name routing algorithm.

\subsection{Training and evaluation data}
We pretrain our models on JFT-4B \citep{zhai2022scaling}, a proprietary dataset that contains more than 4B images, covering 29k classes.
During pretraining, we evaluate the models on two metrics: upstream validation precision-at-1 on JFT-4B, and ImageNet 10-shot accuracy.
The latter is computed by freezing the model weights and replacing the head with a new one that is only trained on a dataset containing 10 images per class from ImageNet-1k \citep{deng2009imagenet}. 
Finally, we provide the accuracy on the validation set of ImageNet-1k after finetuning on the training set of ImageNet-1k (1.3 million images) at 384 resolution.

\subsection{Sparse routing algorithms}

\emph{Tokens Choice}. Every token selects the top-$K$ experts with the highest routing score for the token~\citep{shazeer2017outrageously}.
Increasing $K$ typically leads to better performance at increased computational cost.
Batch Priority Routing (BPR) \citep{riquelme2021scaling} significantly improves the model performance, especially in the case of $K=1$ (\cref{app:additional_results}, \cref{table:bpr_vs_no_bpr}).
Accordingly we use Top-$K$ routing with BPR and $K \in \{1, 2\}$.
We also optimize the number of experts (\cref{app:additional_results}, \cref{fig:ablation_dropping_topk_bpr}).

\emph{Experts Choice}. Alternatively, experts can select the top-$C$ tokens in terms of routing scores \citep{zhou2022mixture}. $C$ is the buffer size, and we set $E \cdot C = c \cdot T$ where $E$ is the number of experts, $T$ is the total number of tokens in the group, and $c$ is the capacity multiplier.
When $c = 1$, all tokens can be processed via the union of experts.
With Experts Choice routing, it is common that some tokens are simultaneously selected by several experts whereas some other tokens are not selected at all.
\Cref{fig:ablation_dropping_topc}, \cref{app:dropping} illustrates this phenomenon.
We experiment with $c=0.5,1,2$.

\subsection{Training Pareto-optimal models}
\label{sec:pareto_plots}
We trained ViT-\{S/8, S/16, S/32, B/16, B/32, L/16, L/32, H/14\} models and their sparse counterparts.
We trained several variants (varying $K$, $C$ and expert number), totalling 106 models.
We trained for 300k steps with batch size 4096, resolution 224, using a reciprocal square root learning rate schedule.

\begin{figure}[tb]
\centering
\begin{subfigure}[b]{0.49\textwidth}
\includegraphics[width=\textwidth]{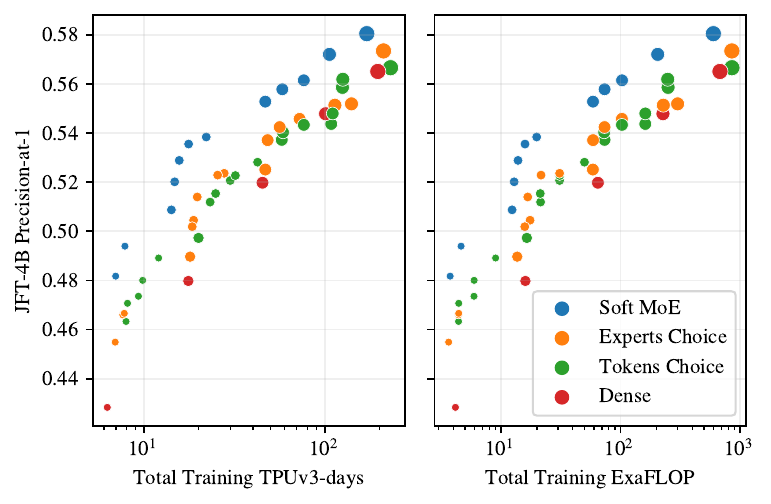}
\caption{JFT-4B Precision-at-1\label{fig:upstream_pareto}}
\end{subfigure}
\hfill
\begin{subfigure}[b]{0.49\textwidth}
\includegraphics[width=\textwidth]{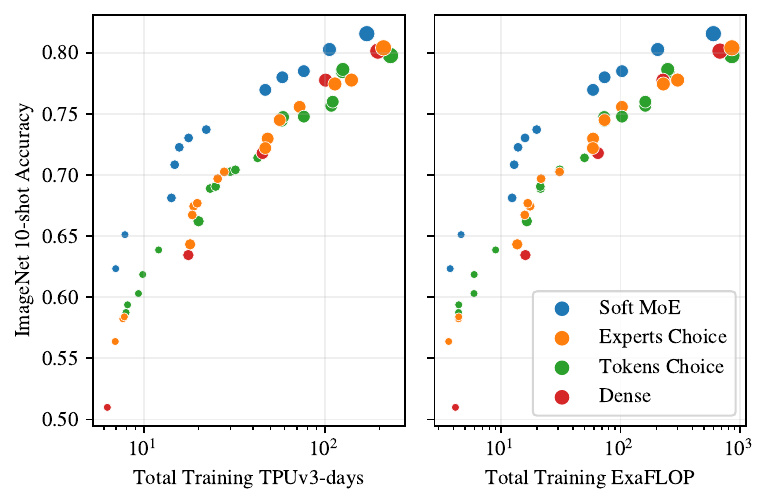}
\caption{ImageNet 10-shot Accuracy\label{fig:imagenet10shot_pareto}}
\end{subfigure}
\caption{
\textbf{Train Pareto frontiers.}
\name dominates both ViTs (dense) and popular MoEs (Experts and Tokens Choice) on the training cost / performance Pareto frontier.
Larger marker sizes indicate larger models, ranging from S/32 to H/14.
Cost is reported in terms of FLOPS and TPU-v3 training time.
Only models on their Pareto frontier are displayed, \cref{app:additional_results} shows all models trained.
\label{fig:small_pareto}}
\end{figure}

\Cref{fig:upstream_pareto,fig:imagenet10shot_pareto} show the results for models in each class that lie on their respective training cost/performance Pareto frontiers.
On both metrics, \name strongly outperforms dense and other sparse approaches for any given FLOPs or time budget.
\cref{table:pareto_runs}, \cref{sec:app_pareto_runs}, lists all the models, with their parameters, performance and costs, which are all displayed in \cref{fig:small_pareto_all}.

\subsection{Long training durations}
\label{sec:long_runs}

We trained a number of models for much longer durations, up to 4M steps.
We trained a number of \names on JFT, following a similar setting to~\citet{zhai2022scaling}.
We replace the last half of the blocks in ViT S/16, B/16, L/16, and H/14 with \name layers with 128 experts, using one slot per expert.
We train models ranging from 1B to 54B parameters.
All models were trained for 4M steps, except for H/14, which was trained for 2M steps for cost reasons.

\begin{figure}[tb]
\centering
\includegraphics[width=\textwidth]{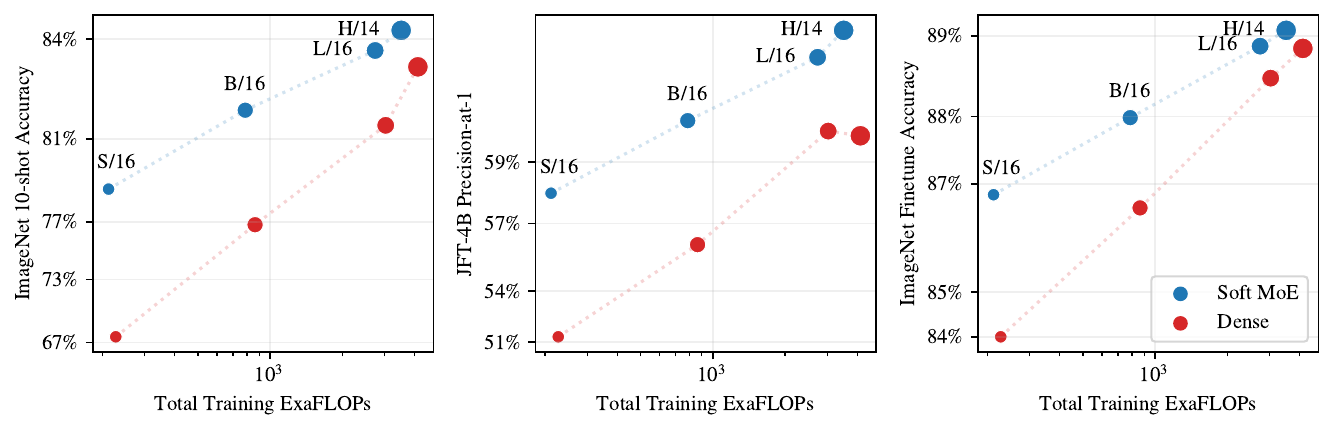}
\caption{%
\textbf{Models with long training durations.}
Models trained for 4M steps (H/14 trained only for 2M steps).
Equivalent model classes (S/16, B/16, etc.) have similar training costs, but \name outperforms ViT on all metrics at a fixed training budget.
\label{fig:softmoe_vs_vit_long_runs}}
\end{figure}

\begin{figure}[tb]
\centering
\includegraphics[width=\textwidth]{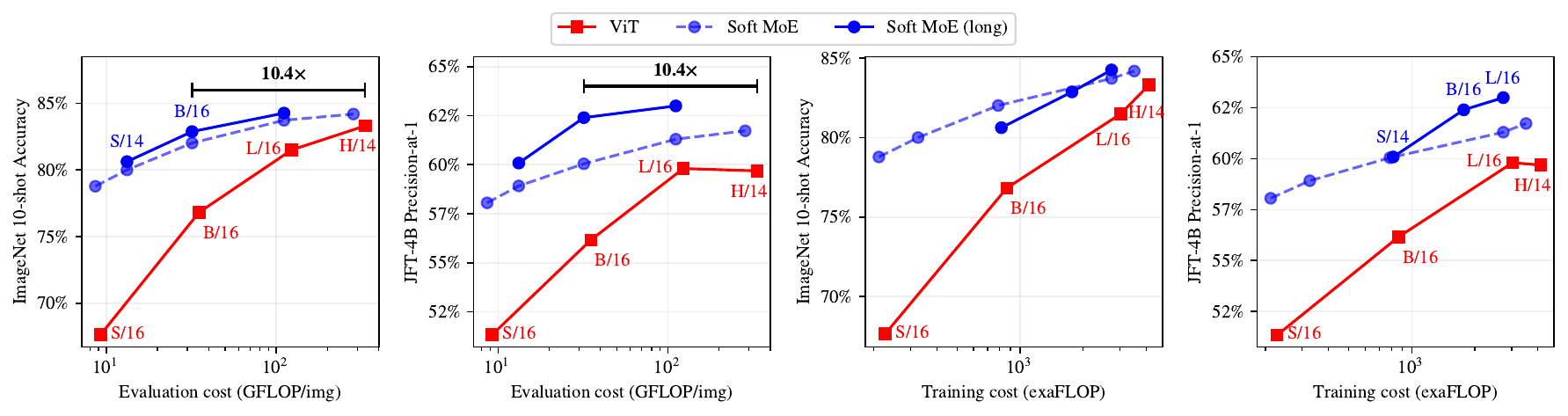}
\includegraphics[width=\textwidth]{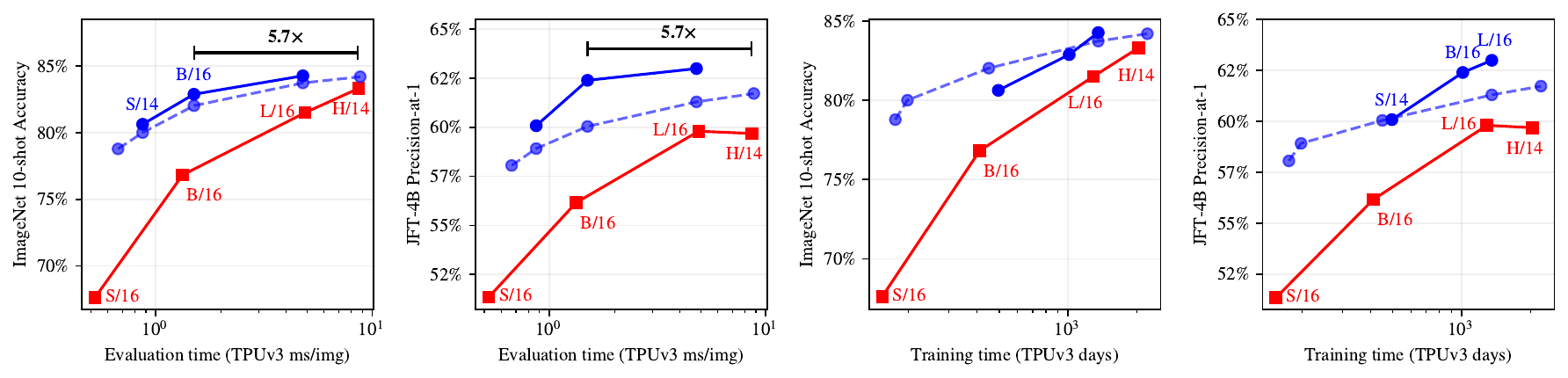}
\caption{
\textbf{Models optimized for inference speed.} Performance of models trained for more steps, thereby optimized for performance at a given inference cost (TPUv3 time or FLOPs). 
\label{fig:long_runs}}
\end{figure}

\Cref{fig:softmoe_vs_vit_long_runs} shows the JFT-4B precision, ImageNet 10-shot accuracy, and the ImageNet finetuning accuracy for \name and ViT versus training cost.
\cref{app:additional_results}, \cref{tab:long_runs_upstream} contains numerical results, and \cref{fig:softmoe_vs_vit_long_runs_time} shows performance versus core-hours, from which the same conclusions can be drawn.
\name substantially outperforms dense ViT models for a given compute budget.
For example, the \name S/16 performs better than ViT B/16 on JFT and 10-shot ImageNet, and it also improves finetuning scores on the full ImageNet data, even though its training (and inference) cost is significantly smaller.
Similarly, \name B/16 outperforms ViT L/16 upstream, and only lags 0.5 behind after finetuning while being 3x faster and requiring almost 4x fewer FLOPs.
Finally, the \name L/16 model outperforms the dense H/14 one while again being around 3x faster in terms of training and inference step time.

We continue training the small backbones up to 9M steps to obtain models of high quality with low inference cost. 
Even after additional (over) training, the overall training time with respect to larger ViT models is similar or smaller.
For these runs, longer cooldowns (linear learning rate decay) works well for \name.
Therefore, we increase the cooldown from 50k steps to 500k steps.

\Cref{fig:long_runs} and \cref{tab:long_runs} present the results.
\name B/16 trained for 1k TPUv3 days matches or outperforms ViT H/14 trained on a similar budget, and is \textbf{10$\times$ cheaper at inference} in FLOPs (32 vs. 334 GFLOPS/img) and \textbf{$\bf >5\times$ cheaper} in wall-clock time (1.5 vs. 8.6 ms/img). 
\name B/16 matches the ViT H/14 model's performance when we double ViT-H/14's training budget (to 2k TPU-days).
\name L/16 \textbf{outperforms all ViT models while being almost 2$\times$ faster at inference} than ViT H/14 (4.8 vs. 8.6 ms/img).

\begin{table}[bt]
\begin{center}
\caption{Models trained for longer durations (cooldown steps in parentheses).
\label{tab:long_runs}}
\vspace{-3mm}
\resizebox{\textwidth}{!}{%
\setlength{\tabcolsep}{2pt} %
\begin{tabular}{lrrrrrrrrr}
\toprule
Model & Params & \multicolumn{1}{c}{Train} & \multicolumn{1}{c}{Train} & \multicolumn{1}{c}{Train} & \multicolumn{1}{c}{Eval} & \multicolumn{1}{c}{Eval}  & \multicolumn{1}{c}{JFT @1} &  \multicolumn{1}{c}{INet} &  \multicolumn{1}{c}{INet} \\
 &  & \multicolumn{1}{c}{steps (cd)} & \multicolumn{1}{c}{TPU-days} & \multicolumn{1}{c}{exaFLOP} & \multicolumn{1}{c}{ms/img} & \multicolumn{1}{c}{GFLOP/img}  & \multicolumn{1}{c}{P@1} & \multicolumn{1}{c}{10shot} & \multicolumn{1}{c}{finetune} \\
\midrule
ViT S/16  &    33M &    4M (50k) &       153.5 &       227.1 &          0.5 &           9.2 &     51.3 &       67.6 & 84.0 \\
ViT B/16  &   108M &    4M (50k) &       410.1 &       864.1 &          1.3 &          35.1 &     56.2 &       76.8 & 86.6 \\
ViT L/16  &   333M &    4M (50k) &      1290.1 &      3025.4 &          4.9 &         122.9 &     59.8 &       81.5 & 88.5 \\
ViT H/14  &   669M &    1M (50k) &      1019.9 &      2060.2 &          8.6 &         334.2 &     58.8 &       82.7 & 88.6 \\
ViT H/14  &   669M &    2M (50k) &      2039.8 &      4120.3 &          8.6 &         334.2 &     59.7 &       83.3 & 88.9 \\
\midrule
\name S/14 256E  &   1.8B &   10M (50k) &       494.7 &       814.2 &          0.9 &          13.2 &     60.1 &       80.6 &   87.5 \\
\name B/16 128E  &   3.7B &   9M (500k) &      1011.4 &      1769.5 &          1.5 &          32.0 &     62.4 &       82.9 &   88.5 \\
\name L/16 128E  &  13.1B &   4M (500k) &      1355.4 &      2734.1 &          4.8 &         111.1 &     63.0 &       84.3 &   89.2 \\
\bottomrule
\end{tabular}}
\end{center}
\end{table}

\subsection{Number of slots and experts}
\label{sec:analysis}

\begin{figure}[tb]
\centering
\includegraphics[width=\textwidth]{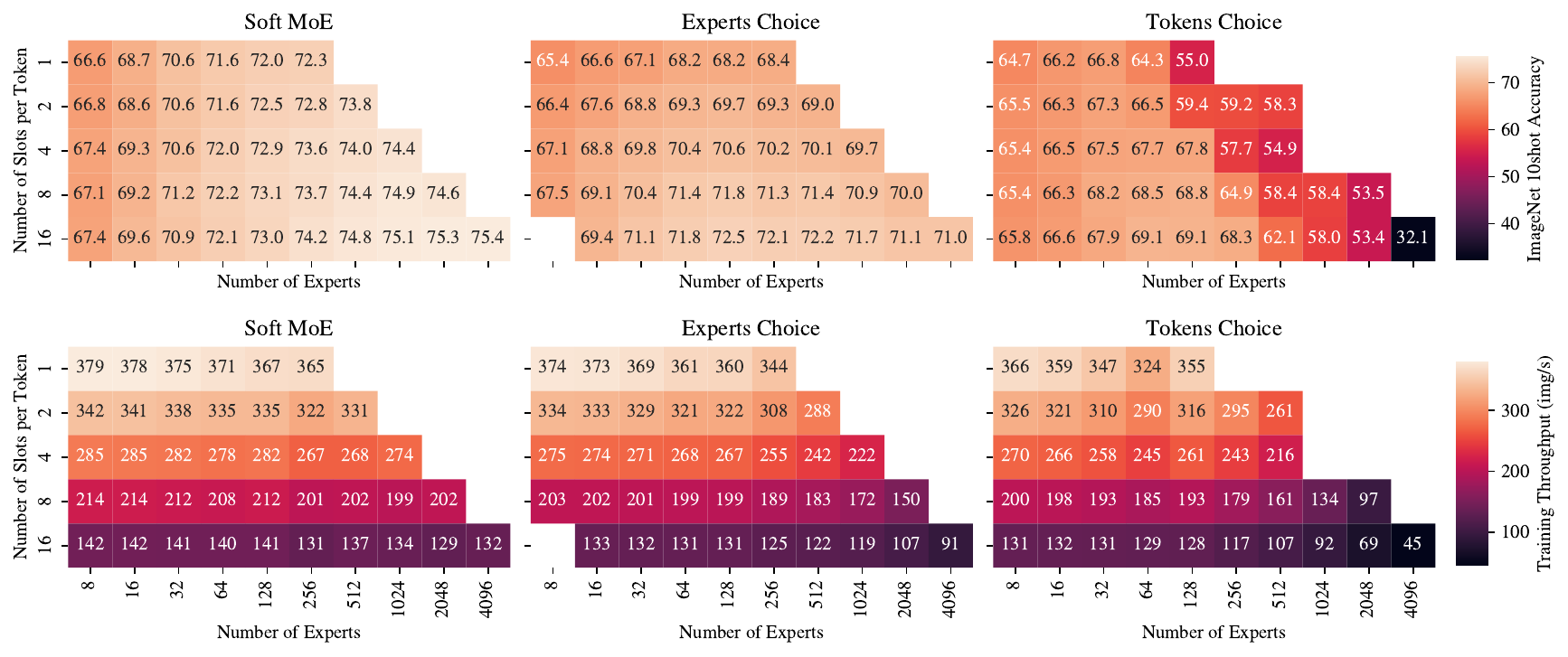}
\caption{
\textbf{Top}: Performance (ImageNet) for MoEs with different number of experts (columns) and slots-per-token / assignments-per-token (rows).
\textbf{Bottom}: Training throughput of the same models.
Across the columns, the number of parameters increases, however, the theoretical cost (FLOPS) for the model (not including routing cost) remains constant.
Descending the rows, the expert layers become more compute intensive as more tokens/slots are processed in the MoE layers.
\label{fig:heatmap}}
\end{figure}

We study the effect of changing the number of slots and experts in the Sparse and Soft MoEs.
\Cref{fig:heatmap} shows the quality and speed of MoEs with different numbers of experts, and numbers of slots per token;
the latter is equivalent to the average number of experts assigned per token for Sparse MoEs.
When varying the number of experts, the models' backbone FLOPs remain constant, so changes in speed are due to routing costs.
When varying the slots-per-expert, the number of tokens processed in the expert layers increases, so the throughput decreases.
First, observe that for \name, the best performing model at each number of slots-per-token is the model with the most experts (i.e. one slot per expert).
For the two Sparse MoEs, there is a point at which training difficulties outweigh the benefits of additional capacity, resulting in the a modest optimum number of experts.
Second, \name's throughput is approximately constant when adding more experts.
However, the Sparse MoEs' throughputs reduce dramatically from 1k experts, see discussion in \cref{sec:soft_moe_vs_sparse}.

\subsection{Ablations}
\label{sec:abaltions}
We study the impact of the components of the \name routing layer by running the following ablations:
\emph{Identity routing}:
Tokens are not mixed: the first token goes to first expert, the second token goes to second expert, etc.
\emph{Uniform Mixing}:
Every slot mixes all input tokens in the same way: by averaging them, both for dispatching and combining.
Expert diversity arises from different initializations of their weights.
\emph{Soft / Uniform}:
We learn token mixing on input to the experts to create the slots (dispatch weights), but we average the expert outputs.
This implies every input token is identically updated before the residual connection.
\emph{Uniform / Soft}.
All slots are filled with a uniform average of the input tokens.
We learn slot mixing of the expert output tokens depending on the input tokens.
\Cref{table:ablation_liquid} shows that having slots is important; Identity and Uniform routing substantially underperform \name, although they do outperform ViT.
Dispatch mixing appears slightly more important than the combine mixing.
See \cref{app:soft_vs_uniform_vs_identity} for additional details.

\begin{table}[t]
\begin{center}
\vspace{-2mm}
\caption{Ablations using \name-S/14 with 256 experts trained for 300k steps.}
\vspace{-3mm}
\label{table:ablation_liquid}
\resizebox{\textwidth}{!}{%
\begin{tabular}{ccccccc}
\toprule
Method & Experts & Mixing & Learned Dispatch & Learned Combine & JFT p@1 & IN/10shot \\
\midrule
\name & \checkmark & \checkmark & \checkmark & \checkmark & 54.3\% & 74.8\% \\
Soft / Uniform & \checkmark & \checkmark & \checkmark &  & 53.6\% & 72.0\% \\
Uniform / Soft & \checkmark & \checkmark &  & \checkmark & 52.6\% & 71.8\% \\
Uniform  & \checkmark & \checkmark &  &  & 51.8\% & 70.0\% \\
Identity & \checkmark &  &  &  & 51.5\% & 69.1\% \\
ViT &  &  & &  & 48.3\% & 62.3\% \\
\bottomrule
\end{tabular}}
\end{center}
\vspace{-3mm}
\end{table}

\section{Contrastive learning}
\label{sec:contrastive_experiments}
We test whether the \name's representations are better for other tasks.
For this, we try image-text contrastive learning.
Following \citet{zhai2022lit}, the image tower is pre-trained on image classification, and then frozen while training the text encoder on a dataset of image-text pairs.
We re-use the models trained on JFT in the previous section and compare their performance zero-shot on downstream datasets.
For contrastive learning we train on WebLI~\citep{chen2022pali}, a proprietary dataset consisting of 10B images and alt-texts.
The image encoder is frozen, while the text encoder is trained from scratch.

\Cref{table:lit_0shot_long_training} shows the results.
Overall, the benefits we observed on image classification are also in this setting.
For instance, \name-L/16 outperforms ViT-L/16 by more than 1\% and 2\% on ImageNet and Cifar-100 zero-shot, respectively.
However, the improvement on COCO retrieval are modest, and likely reflects the poor alignment between features learned on closed-vocabulary JFT and this open-vocabulary task.

\begin{table}[t]
\begin{center}
\caption{LIT-style evaluation with a ViT-g text tower trained for 18B input images ($\sim 5$ epochs).}
\vspace{-3mm}
\label{table:lit_0shot_long_training}
\setlength{\tabcolsep}{2pt} %
\begin{tabular}{lrrrrrr}
\toprule
Model & Experts & IN/0shot & Cifar100/0shot & Pet/0shot & Coco Img2Text & Coco Text2Img \\
\midrule
ViT-S/16 & -- & 74.2\% & 56.6\% & 94.8\% & 53.6\% & 37.0\% \\
\name-S/16 & 128 & 81.2\% & 67.2\% & 96.6\% & 56.0\% & 39.0\% \\
\name-S/14 & 256 & 82.0\% & 75.1\% & 97.1\% & 56.5\% & 39.4\% \\
\midrule
ViT-B/16 & -- & 79.6\% & 71.0\% & 96.4\% & 58.2\% & 41.5\% \\
\name-B/16 & 128 & 82.5\% & 74.4\% & 97.6\% & 58.3\% & 41.6\% \\
\midrule
ViT-L/16 & -- & 82.7\% & 77.5\% & 97.1\% & 60.7\% & 43.3\% \\
\name-L/16 & 128 & 83.8\% & 79.9\% & 97.3\% & 60.9\% & 43.4\% \\
Souped \name-L/16 & 128 & 84.3\% & 81.3\% & 97.2\% & 61.1\% & 44.5\% \\
\midrule
ViT-H/14 & -- & 83.8\% & 84.7\% & 97.5\% & 62.7\% & 45.2\% \\     %
\name-H/14 & 256 & 84.6\% & 86.3\% & 97.4\% & 61.0\% & 44.8\% \\   %
\bottomrule
\end{tabular}
\end{center}
\end{table}

Finally, in \cref{app:additional_results_laion} we show that \name{}s also surpass vanilla ViT and the Experts Choice router when trained from scratch on the publicly available LAION-400M \citep{schuhmann2021laion}. With this pretraining, \name{}s also benefit from data augmentation, but neither ViT nor Experts Choice seem to benefit from it, which is consistent with our observation in \cref{sec:analysis}, that \name{}s make a better use of additional expert parameters.

\section{Related Work}
\label{sec:soft_moe_vs_others}
Many existing works \emph{merge}, \emph{mix} or \emph{fuse} input tokens to reduce the input sequence length \citep{jaegle2021perceiver,ryoo2021tokenlearner,renggli2022learning,Wang_2022_CVPR}, typically using attention-like weighted averages with fixed keys, to try to alleviate the quadratic cost of self-attention with respect to the sequence length. Although our dispatch and combine weights are computed in a similar fashion to these approaches, our goal is not to reduce the sequence length (while it is possible), and we actually recover the original sequence length after weighting the experts' outputs with the \emph{combine weights}, at the end of each \name layer.

Multi-headed attention also shows some similarities with \name, beyond the use of softmax in weighted averages:
the $h$ different \emph{heads} can be interpreted as different (linear) experts.
The distinction is that, if $m$ is the sequence length and each input token has dimensionality $d$, each of the $h$ heads processes $m$ vectors of size ${d}/{h}$.
The $m$ resulting vectors are combined using different weights for each of the $m'$ output tokens (i.e. the attention weights), on each head independently, and then the resulting $(d/h)$-dimensional vectors from each head are concatenated into one of dimension $d$. Our experts are non-linear and combine vectors of size $d$, at the \emph{input and output} of such experts.

Other MoE works use a weighted combination of the experts parameters, rather than doing a sparse routing of the examples \citep{yang2019condconv,tian2020conditional,muqeeth2023soft}. These approaches are also fully differentiable, but they can have a higher cost, since 1) they must average the parameters of the experts, which can become a time and/or memory bottleneck when experts with many parameters are used; and 2) they cannot take advantage of vectorized operations as broadly as Soft (and Sparse) MoEs, since \emph{every input uses a different weighted combination of the parameters}. We recommend the ``computational cost'' discussion in \citet{muqeeth2023soft}.

\section{Current limitations}
\label{sec:limitations}

\paragraph{Auto-regressive decoding}
One of the key aspects of \name consists in learning the merging of all tokens in the input. This makes the use of \names in auto-regressive decoders difficult, since causality between past and future tokens has to be preserved during training. Although causal masks used in attention layers could be used, one must be careful to not introduce any correlation between token and slot \emph{indices}, since this may bias which token indices each expert is trained on. The use of \name in auto-regressive decoders is a promising research avenue that we leave for future work.
\vspace{-2mm}

\paragraph{Lazy experts \& memory consumption}
We show in \cref{sec:classification_experiments} that one slot per expert tends to be the optimal choice.
In other words, rather than feeding one expert with two slots, it is more effective to use two experts with one slot each.
We hypothesize slots that use the same expert tend to align and provide small informational gains, and a expert may lack the flexibility to accommodate very different slot projections.
We show this in \Cref{app:slot_correlation}.
Consequently, \name can leverage a large number of experts and---while its cost is still similar to the dense backbone---the memory requirements of the model can grow large. 

\clearpage

\bibliography{main}
\bibliographystyle{plainnat}

\clearpage
\appendix
\section{Soft vs. Uniform vs. Identity dispatch and combine weights}
\label{app:soft_vs_uniform_vs_identity}

In this section, we compare \name (i.e.\ the algorithm that uses the dispatch and combine weights computed by \name in \cref{eq:dispatch_weights_def} and \cref{eq:combine_weights_def}) with different ``fixed routing'' alternatives, where neither the expert selected nor the weight of the convex combinations depend on the \emph{content} of the tokens.

We consider the following simple modifications of \name:

\textbf{Identity}. The first token in the sequence is processed by the first expert, the second token by the second expert, and so on in a round robin fashion. When the sequence length is the same as the number of slots and experts, this is equivalent to replacing the matrix $\mD$ in \cref{eq:dispatch_weights_def} (resp. $\mC$ in \cref{eq:combine_weights_def}) with an identity matrix.

\textbf{Uniform}. Every input slot is filled with a uniform average of all input tokens, and every output token is a uniform average of all output slots. This is equivalent to replacing the matrix $\mD$ from \cref{eq:dispatch_weights_def} with values $\frac{1}{m}$ in all elements, and a matrix $\mC$ from \cref{eq:combine_weights_def} with values $\frac{1}{n p}$ in all elements.
We randomly and independently initialize every expert.

\textbf{Uniform / Soft}. Every input slot is filled with a uniform average of all input tokens, but we keep the definition of $\mC$ from \cref{eq:combine_weights_def}.

\textbf{Soft / Uniform}. Every output token is a uniform average of all output slots, but we keep the definition of $\mD$ in \cref{eq:dispatch_weights_def}.

\begin{figure}[hb]
\centering
\includegraphics[width=\textwidth]{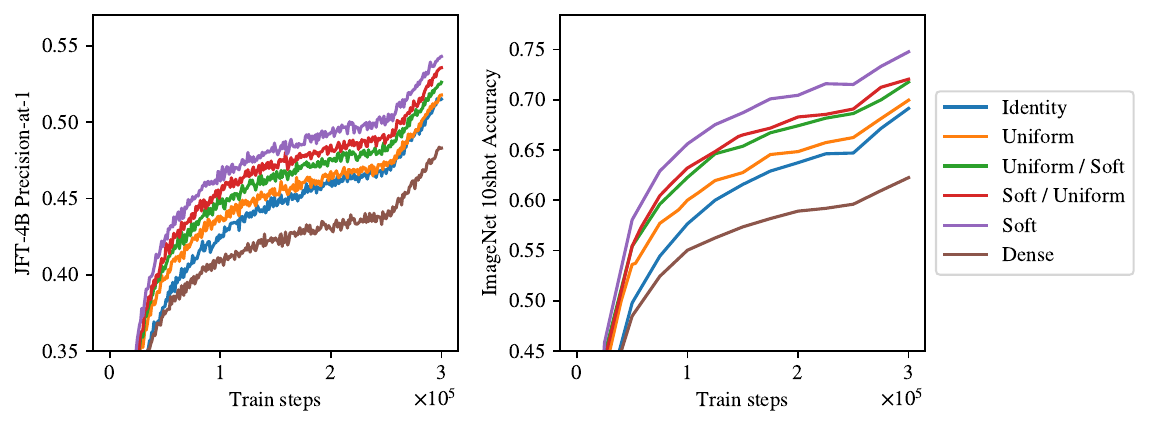}
\caption{%
\textbf{\name compared against different ``fixed routing'' strategies}. \emph{Identity} processes the $i$-th token with the $i$-th expert; \emph{Uniform} replaces both the dispatch and combine matrices with uniform averages; \emph{Uniform / Soft} replaces the dispatch weights with a uniform average, but the combine weights are computed as in \name; \emph{Soft / Uniform} does the opposite replacement; and \emph{Soft} uses the algorithm we present in \cref{sec:softmoe}.
\label{fig:softmoe_vs_uniform_vs_identity}}
\end{figure}

\Cref{fig:softmoe_vs_uniform_vs_identity} and \cref{table:ablation_liquid} shows the results from this experiment, training a S/14 backbone model with MoEs on the last 6 layers. Since the sequence length is 256, we choose 256 experts and slots (i.e. 1 slot per expert), so that the matrices $\mD$ and $\mC$ are squared. As shown in the figure, \name is far better than all the other alternatives.
For context, we also add the dense ViT S/14 to the comparison.

\section{Token Dropping}
\label{app:dropping}

In this appendix, we briefly explore token dropping for the Experts Choose and Tokens Choose algorithms.
For Tokens Choose, each token selects $K$ experts. When experts are full, some tokens assigned to that expert will not be processed. A token is ``dropped'' when none of its choices go through, and no expert at all processes the token.
Expert Choose algorithms lead to an uneven amount of processing per token: some input tokens are selected by many experts, while some others are not selected by any.
We usually define the number of tokens to be processed by each expert in a way that the combined capacity of all experts corresponds to the number of input tokens (or a multiple $C$ of them).
If we use a multiplier $C$ higher than one (say, 2x or 3x), the amount of dropping will decrease but we will pay an increased computational cost.
Thus, we mainly explore the $K=1$ and $C=1$ setup, where there is no slack in the buffers.

In all cases to follow we see a common trend: fixing everything constant, increasing the number of experts leads to more and more dropping both in Experts Choose and Tokens Choose.

\begin{figure}[tb]
\centering
\includegraphics[width=\textwidth]{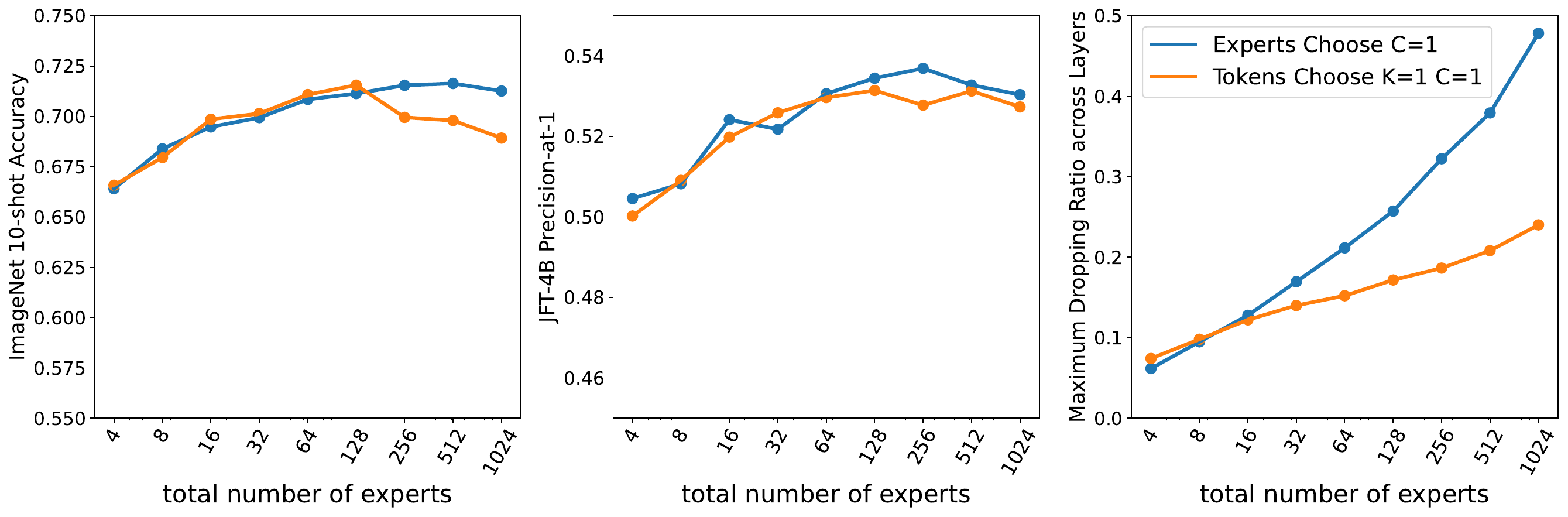}
\caption{S/14. Performance and amount of token dropping for increasing experts for Experts Choose ($C=1$) and Tokens Choose ($K=1$ and $C=1$).}
\label{fig:ablation_dropping_topk_topc_1}
\end{figure}

\Cref{fig:ablation_dropping_topk_topc_1} compares Experts Choose and Tokens Choose with the same multiplier $C=1$.
This is the cheapest setup where every token \emph{could} be assigned to an expert with balanced routing.
We see that in both cases the amount of dropping quickly grows with the number of experts.
Moreover, even though Experts Choose has higher levels of dropping (especially for large number of experts), it is still more performant than Tokens Choose.
Note there is a fundamental difference: when Tokens Choose drops a token, the model wastes that amount of potential compute.
On the other hand, for Experts Choose dropping just means some other token got that spot in the expert buffer, thus the model just transferred compute from one unlucky token to another lucky one.

In this setup, for a small number of experts (16-32) it is common to observe a $\sim15\%$ rate of dropping.
On the other hand, we also experimented with a large number of experts (100-1000) where each expert selects very few tokens.
In this case, the dropping rate for Experts Choose can grow above 40-50\% in some layers: most experts select the very same tokens.
Tokens Choose seems to completely drop up to $\sim$25\% of the tokens.

\begin{figure}[tb]
\centering
\includegraphics[width=\textwidth]{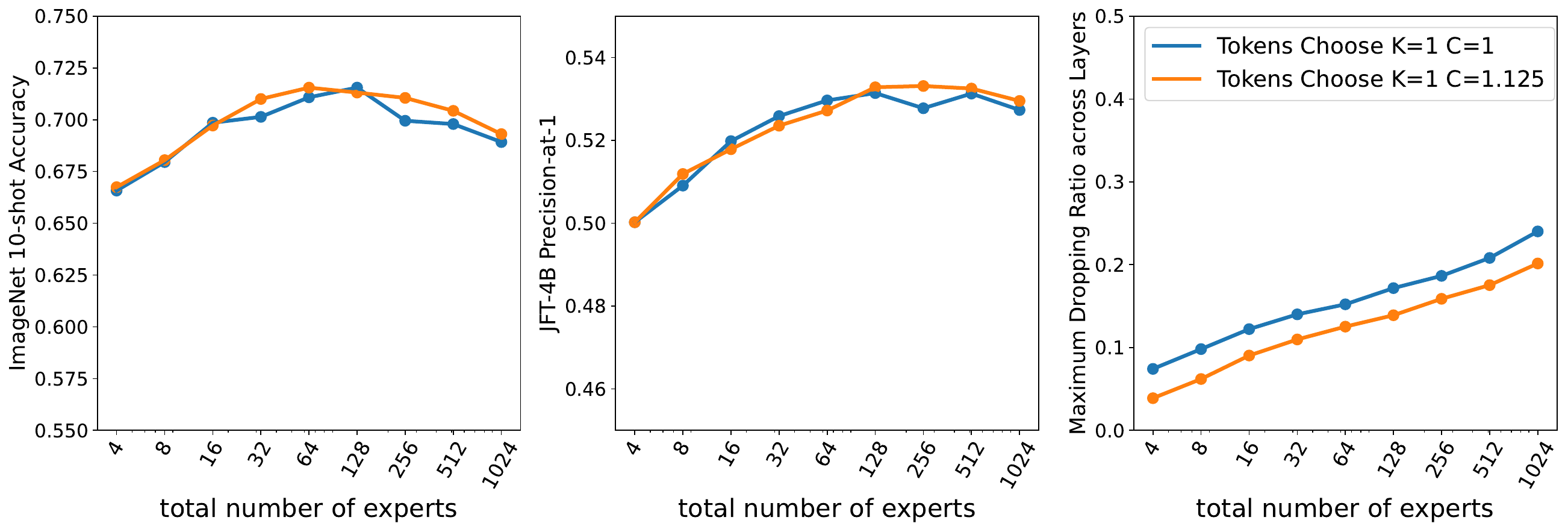}
\caption{S/14. Performance and amount of token dropping for increasing experts for Tokens Choose with tight buffers ($K=1$ and $C=1$) and some amount of buffer slack ($K=1$ and $C=1.125$).}
\label{fig:ablation_dropping_topk}
\end{figure}

\begin{figure}[tb]
\centering
\includegraphics[width=\textwidth]{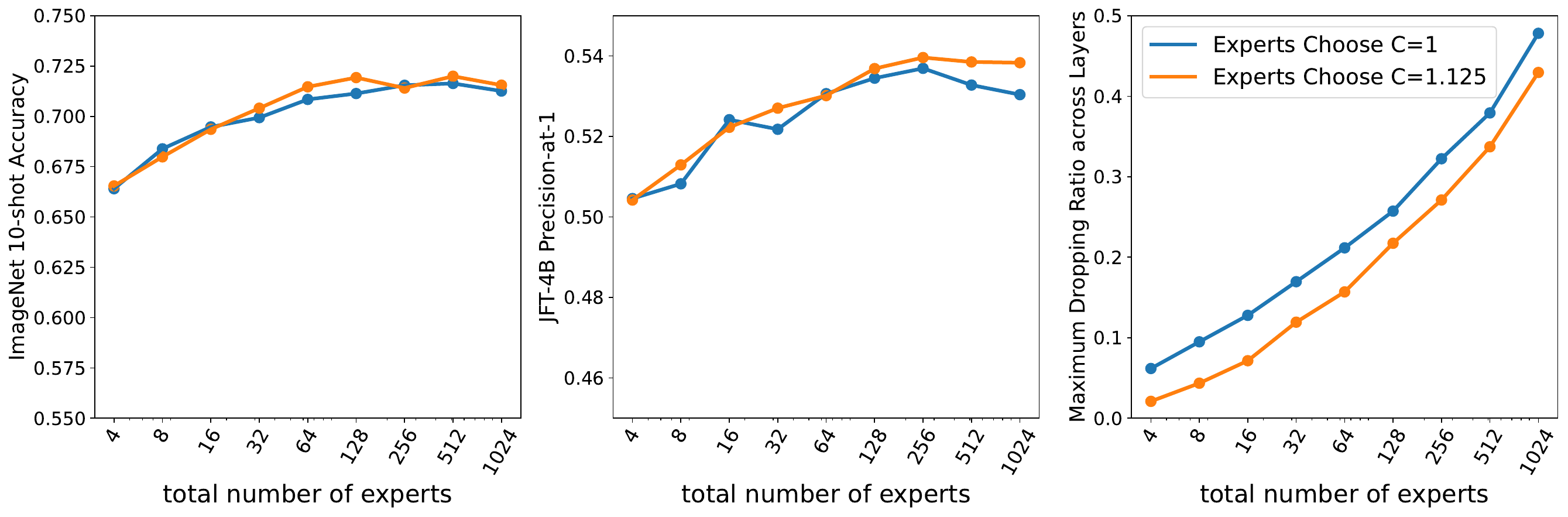}
\caption{S/14. Performance and amount of token dropping for increasing experts for Experts Choose with tight buffers ($C=1$) and slightly larger buffers ($C=1.125$).}
\label{fig:ablation_dropping_topc}
\end{figure}

In \cref{fig:ablation_dropping_topc,fig:ablation_dropping_topk} we study how much a little bit of buffer slack ($C=1.125$) can help in terms of performance and dropping to Experts Choose and Tokens Choose, respectively.
Both plots are similar: the amount of dropping goes down around $\sim$5\% and performance slightly increases when the number of experts is large.
Note that the step time also increases in these cases.

\begin{figure}[tb]
\centering
\includegraphics[width=\textwidth]{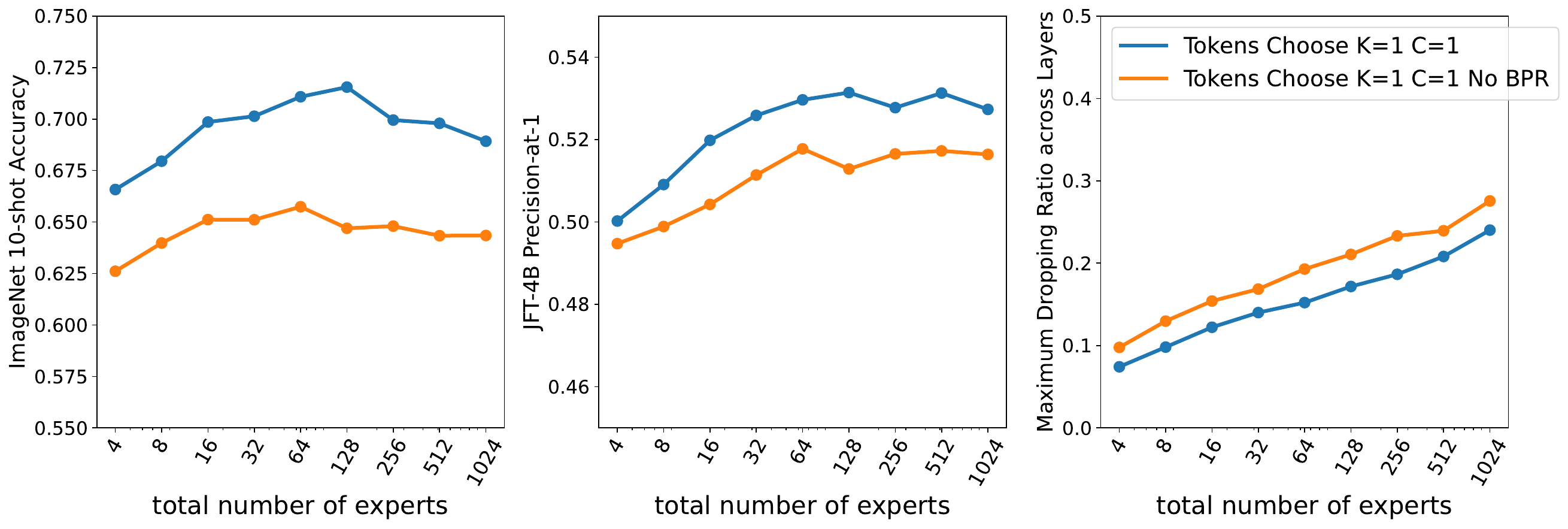}
\caption{S/14. Performance and amount of token dropping for increasing experts with and without BPR for Tokens Choose.}
\label{fig:ablation_dropping_topk_bpr}
\end{figure}

Finally, \cref{fig:ablation_dropping_topk_bpr} shows the effect of Batch Priority Routing \citep{riquelme2021scaling} for Tokens Choose. By smartly selecting which tokens to drop we do not only uniformly reduce the amount of dropping, but we significantly bump up performance.

\section{\name Increasing Slots}
\label{app_increasing_slots}

\begin{figure}[htb]
\centering
\includegraphics[width=\textwidth]{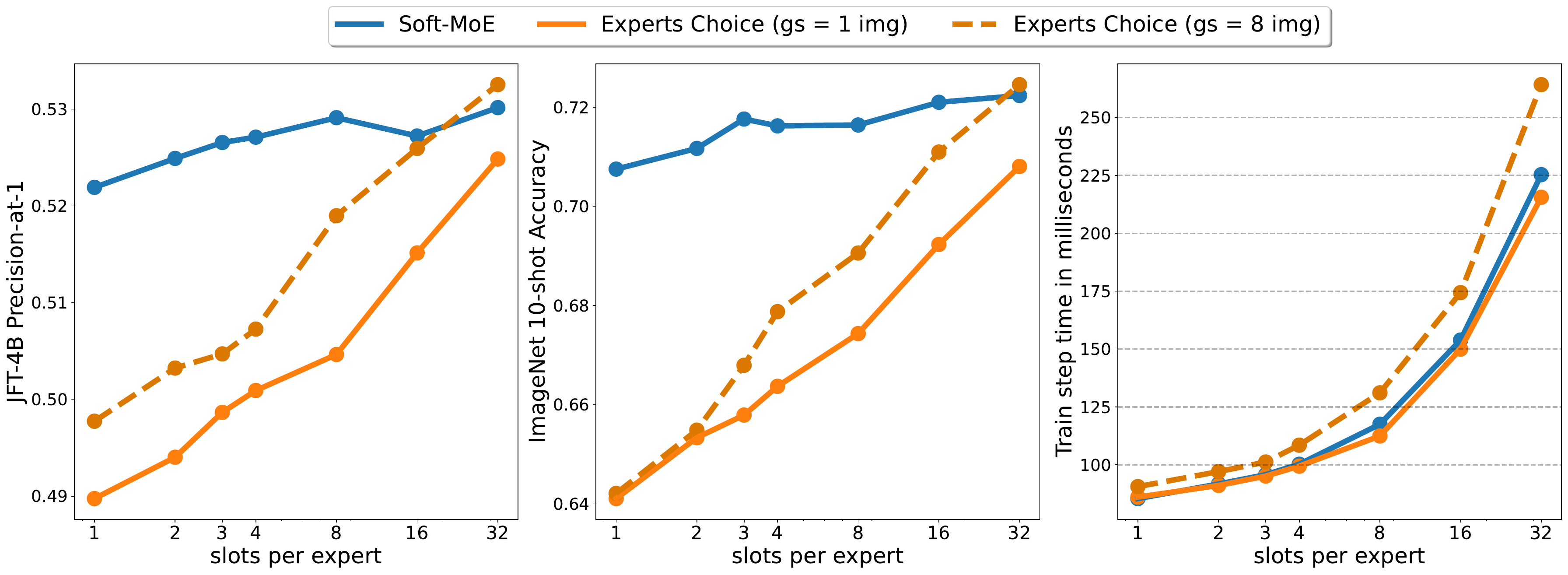}
\caption{
\textbf{Performance (left, center) and step time (right) of models with 32 experts, but increased slots, all trained for the same number of steps (300k).}
Increasing the number of slots per expert only increases performance of \name a small amount, while increasing cost substantially.
\label{fig:ablation_increasing_slots_per_expert}}
\end{figure}

In this section we explore the following question: for a fixed number of experts, how much does \name routing benefit from having additional slots per expert?
\Cref{fig:ablation_increasing_slots_per_expert} shows results for \name S/16 with 32 experts.
We also show Experts Choice with group sizes of one and eight images.
When increasing the number of slots, the performance grows only modestly, while cost increases quickly.
Experts Choice benefits much more from increased slots, catching up at a large group size, but at a very large cost.

\section{Sparse Layers Placement}
\label{app_sparse_placement}

\name does unlock the effective use of a large number of experts.
An important design choice for sparse models is the number and location of sparse layers, together with the number of experts per layer.
Unfortunately, the large number of degrees of freedom in these choices has usually made thorough ablations and optimization unfeasible.
In this section, we provide the results of a simple experiment that can help better design the configuration of sparse models.
We fix a total number of experts ($E=512$) with one slot per expert, thus leading to matched number of parameters (note in this case FLOPs may vary greatly depending on the number of sparse layers).
Then, for an S/16 backbone architecture, we distribute those experts in various ways (all in one layer, half of them in two layers, etc) and compare their performance after 300k training steps.
\Cref{table:expert_placement} shows the results.
Again, we observe that a number of experts close to the number of input tokens (there are 196 tokens, given the 16x16 patch size for 224x224 images) split over the last few layers works best.
Moreover, note these models are indeed cheaper than those in the comparison with 512 or 256 experts per layer.
\Cref{table:expert_placement_topk} offers results for Tokens Choose routing with $K=1$ and BPR \cite{riquelme2021scaling}.
In this case, all algorithms use a comparable FLOPs count (ignoring slightly increasing routing costs with more experts).
Results are essentially similar, thus suggesting optimal expert placement (including expert count and location) may not strongly depend on the routing algorithm.

\begin{table}[htb]
\begin{center}
\caption{Expert placing ablation with a \name S/16 with 12 layers (indexed from 0 to 11).}
\label{table:expert_placement}
\begin{tabular}{ ccccccc }
\toprule
Sparse Layers & Experts per Layer & Total Experts & IN/10shot & JFT prec\@1 \\
\midrule
11 & 512 & 512 & 70.0\% & 51.5\% \\ 
10 & 512 & 512 & 70.1\% & 52.0\% \\ 
\midrule
10, 11 & 256 & 512 & 71.7\% & 52.2\% \\ 
5, 11 & 256 & 512 & 70.4\% & 52.1\% \\ 
\midrule
8, 9, 10, 11 & 128 & 512 & \textbf{72.8\%} & \textbf{53.2\%} \\ 
2, 5, 8, 11 & 128 & 512 & 71.1\% & 52.5\% \\ 
\midrule
4:11 & 64 & 512 & \textbf{72.1\%} & \textbf{53.1\%} \\ 
1:4, 8:11 & 64 & 512 & 70.5\% & 52.1\% \\ 
\bottomrule
\end{tabular}
\end{center}
\end{table}

\begin{table}[htb]
\begin{center}
\caption{Expert placing ablation with a V-MoE S/16 Tokens Choose $K=1$ with 12 layers (indexed as 0:11).}
\label{table:expert_placement_topk}
\begin{tabular}{ ccccccc }
\toprule
Sparse Layers & Experts per Layer & Total Experts & IN/10shot & JFT prec\@1 \\
\midrule
11 & 512 & 512 & 64.4\% & 50.1\% \\ 
10 & 512 & 512 & 67.2\% & 51.9\% \\ 
\midrule
10, 11 & 256 & 512 & 68.6\% & 51.3\% \\ 
5, 11 & 256 & 512 & 65.3\% & 50.6\% \\ 
\midrule
8, 9, 10, 11 & 128 & 512 & \textbf{69.1\%} & \textbf{52.3\%} \\ 
2, 5, 8, 11 & 128 & 512 & 67.3\% & 51.1\% \\ 
\midrule
4:11 & 64 & 512 & \textbf{69.9\%} & \textbf{52.2\%} \\ 
1:4, 8:11 & 64 & 512 & 68.0\% & 51.2\% \\ 
\bottomrule
\end{tabular}
\end{center}
\end{table}

\begin{table}[htb]
\begin{center}
\caption{Expert placing ablation with a V-MoE S/16 Experts Choose $C=1$ with 12 layers (indexed as 0:11).}
\label{table:expert_placement_topc}
\begin{tabular}{ ccccccc }
\toprule
Sparse Layers & Experts per Layer & Total Experts & IN/10shot & JFT prec\@1 \\
\midrule
11 & 512 & 512 & 65.3\% & 50.3\% \\ 
10 & 512 & 512 & 66.5\% & 51.7\% \\ 
\midrule
10, 11 & 256 & 512 & 68.8\% & 51.8\% \\ 
5, 11 & 256 & 512 & 65.9\% & 51.1\% \\ 
\midrule
8, 9, 10, 11 & 128 & 512 & \textbf{69.4\%} & \textbf{52.2\%} \\ 
2, 5, 8, 11 & 128 & 512 & 68.0\% & 51.7\% \\ 
\midrule
4:11 & 64 & 512 & \textbf{69.0\%} & \textbf{52.2\%} \\ 
1:4, 8:11 & 64 & 512 & 67.4\% & 51.1\% \\ 
\bottomrule
\end{tabular}
\end{center}
\end{table}

\clearpage
\section{The collapse of softmax layers applied after layer normalization}\label{sec:layernorm_problems}

\subsection{Theoretical analysis}
A softmax layer with parameters $\Theta \in \R^{n \times d}$ transforms a vector $x \in R^d$ into the vector $\text{softmax}(\Theta x) \in \R^n$, with elements:
\begin{equation}
\text{softmax}(\Theta x)_i = \frac{\exp((\Theta x)_i)}{\sum_{j=1}^n \exp((\Theta x)_j)} =
\frac{\exp(\sum_{k=1}^d \theta_{ik}x_k) }{\sum_{j=1}^n \exp(\sum_{k=1}^d \theta_{jk}x_k)}
\end{equation}

Layer normalization applies the following operation on $x \in \R^d$.
\begin{equation}
\text{LN}(x)_i = \alpha_i \frac{x_i - \mu(x)}{\sigma(x)} + \beta_i;
~~ \text{ where } ~ \mu(x) = \frac{1}{d} \sum_{i=1}^d x_i ~ \text{ and } ~ \sigma(x) = \sqrt{\frac{1}{d}  \sum_{i=1}^d (x_i - \mu(x_i))^2}
\end{equation}

Notice that $\text{LN}(x) = \text{LN}(x - \mu(x))$, thus we can rewrite LayerNorm with respect to the centered vector $\tilde{x} = x - \mu(x)$, and the centered vector scaled to have unit norm $\hat{x}_i = \frac{\tilde{x}_i}{\|\tilde{x}\|}$: 
\begin{equation}
\text{LN}(\tilde{x})_i = \alpha_i \frac{\tilde{x}_i}{\sqrt{\frac{1}{d} \sum_{j=1}^{d} \tilde{x}_j^2}} + \beta_i = \sqrt{d} \alpha_i \frac{\tilde{x}_i}{\|\tilde{x}\|} + \beta_i = \sqrt{d} \alpha_i \hat{x}_i  + \beta_i
\end{equation}

When a softmax layer is applied to the outputs of layer normalization, the outputs of the softmax are given by the equation:
\begin{equation}
\text{softmax}(\Theta \text{LN}(x))_i = 
\frac{\exp(\sum_{k=1}^d \theta_{ik}(\sqrt{d} \alpha_k \hat{x}_k  + \beta_k)) }{\sum_{j=1}^n \exp(\sum_{k=1}^d \theta_{jk}(\sqrt{d} \alpha_k \hat{x}_k  + \beta_k))}
\end{equation}

By setting $\vartheta_{i} = \sum_{k=1}^d \theta_{ik} \alpha_k \hat{x}_k$, and $\delta_i = \sum_{k=1}^d \theta_{ik} \beta_k$, the previous equation can be rewritten as:
\begin{equation}
\text{softmax}(\Theta \text{LN}(x))_i = 
\frac{\exp(\sqrt{d} \vartheta_i  + \delta_i) }{\sum_{j=1}^n \exp(\sqrt{d} \vartheta_j  + \delta_j)}
\end{equation}

Define $m = \max_{i \in [n]} \sqrt{d} \vartheta_i - \delta_i$, $M = \{i \in [n] : \sqrt{d} \vartheta_i - \delta_i = m\}$. Then, the following equality holds:
\begin{equation}
\text{softmax}(\Theta \text{LN}(x))_i = 
\frac{\exp(\sqrt{d} \vartheta_i  + \delta_i - m) }{\sum_{j=1}^n \exp(\sqrt{d} \vartheta_j  + \delta_j - m)}
\end{equation}

Given that $\lim_{d\rightarrow \infty} \exp(\sqrt{d} \vartheta_i  + \delta_i - m) = \begin{cases}1 : i \in M\\ 0 : i \notin M\end{cases}$ the output of the softmax tends to:
\begin{equation}
\lim_{d \rightarrow \infty} \text{softmax}(\Theta \text{LN}(x))_i =
\begin{cases}
\frac{1}{|M|} & i \in M\\
0 & i \notin M
\end{cases}
\end{equation}

In particular, when the maximum is only achieved by one of the components (i.e. $|M| = 1$), the softmax collapses to a one-hot vector (a vector with all elements equal to 0 except for one).

\subsection{Empirical analysis}
The previous theoretical analysis assumes that the parameters of the softmax layer are constants, or more specifically that they do not depend on $d$.
One might argue that using modern parameter initialization techniques, which take into account $\frac{1}{\sqrt{d}}$ in the standard deviation of the initialization \cite{glorot2010understanding,he2015delving,klambauer2017self}, might fix this issue. We found that they don't (in particular, we use the initialization from \cite{glorot2010understanding}).

\begin{figure}[bt]
\centering
\begin{subfigure}[b]{\textwidth}
\centering
\includegraphics[width=\textwidth]{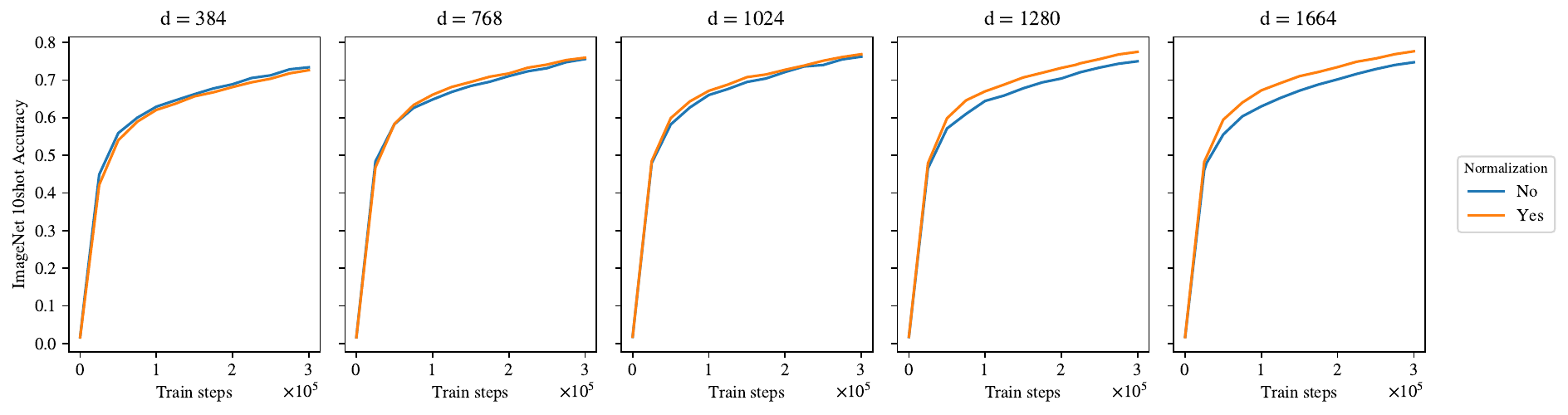}
\caption{ImageNet 10shot accuracy.}
\end{subfigure}
\begin{subfigure}[b]{\textwidth}
\centering
\includegraphics[width=\textwidth]{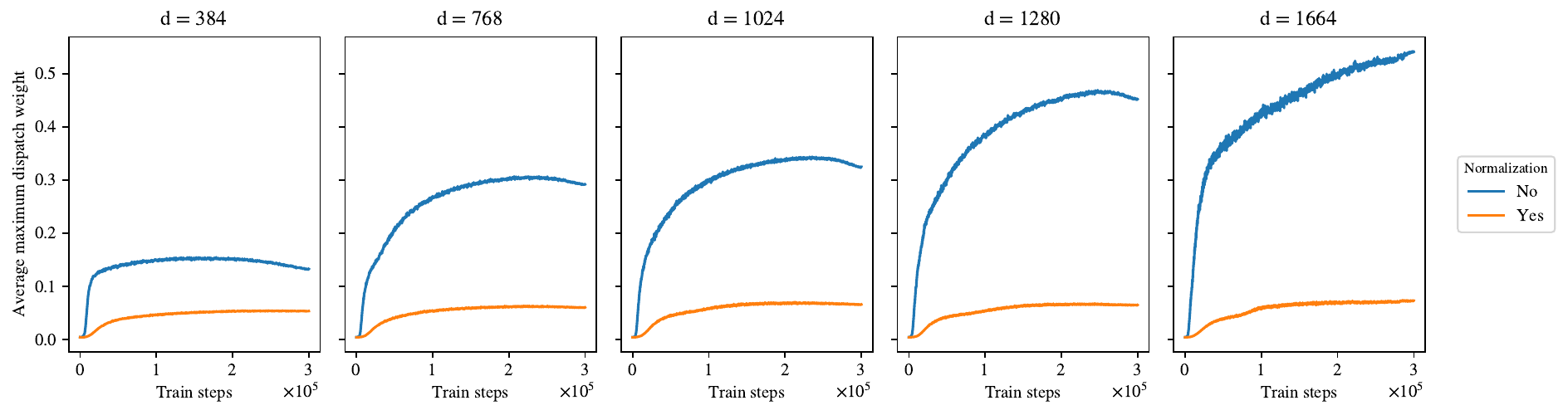}
\caption{Average value of the maximum dispatch weight per slot.}
\end{subfigure}
\begin{subfigure}[b]{\textwidth}
\centering
\includegraphics[width=\textwidth]{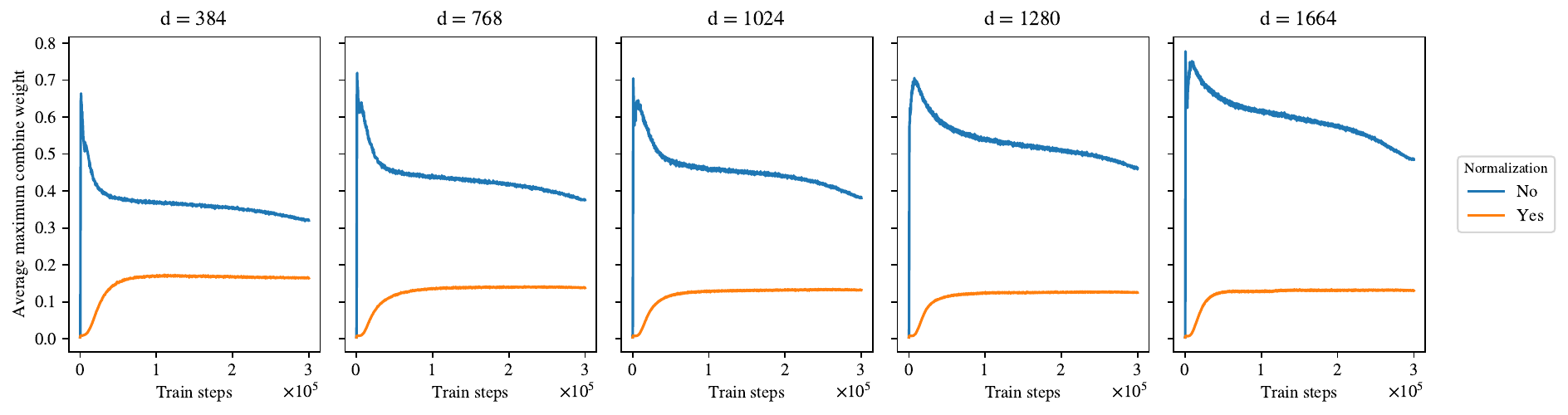}
\caption{Average value of the maximum combine weight per token.}
\end{subfigure}
\caption{%
Training plots of the ImageNet 10shot accuracy (top), the average value of the maximum dispatch weight per slot (middle) and the average value of the maximum combine weight per token (bottom) for different model dimensions $d$. Observe that maximum values of the combine and (especially) the dispatch weights grow as the model dimension grows during training, as our theoretical analysis predicted. Although the ImageNet 10shot accuracy is similar for small model dimensions, applying the softmax layer directly on the output of layer normalization, without any further re-normalization, hurts the accuracy as the model dimension $d$ grows. By normalizing the inputs to the softmax as suggested in \cref{sec:implementation} improves the performance for large values of $d$.
\label{fig:layernorm_issues_model_dim}}
\end{figure}

\Cref{fig:layernorm_issues_model_dim} shows different metric curves during the training of a small SoftMoE model with different model dimensions. The model dimensions are those corresponding to different standard backbones: S (384), B (768), L (1024), H (1280) and G (1664). The rest of the architecture parameters are fixed: 6 layers (3 dense layers followed by 3 MoE layers with 256 experts), 14x14 patches, and a MLP dimension of 1536. As the model dimension $d$ increases, the figure shows that, if the inputs to the softmax in the SoftMoE layers are not normalized, the average maximum values of the dispatch and combine weights tend to grow (especially the former). When $d$ is big enough, the ImageNet 10shot accuracy is significantly worse than that achieved by properly normalizing the inputs.

\begin{figure}[bt]
\centering
\begin{subfigure}[b]{\textwidth}
\includegraphics[width=\textwidth]{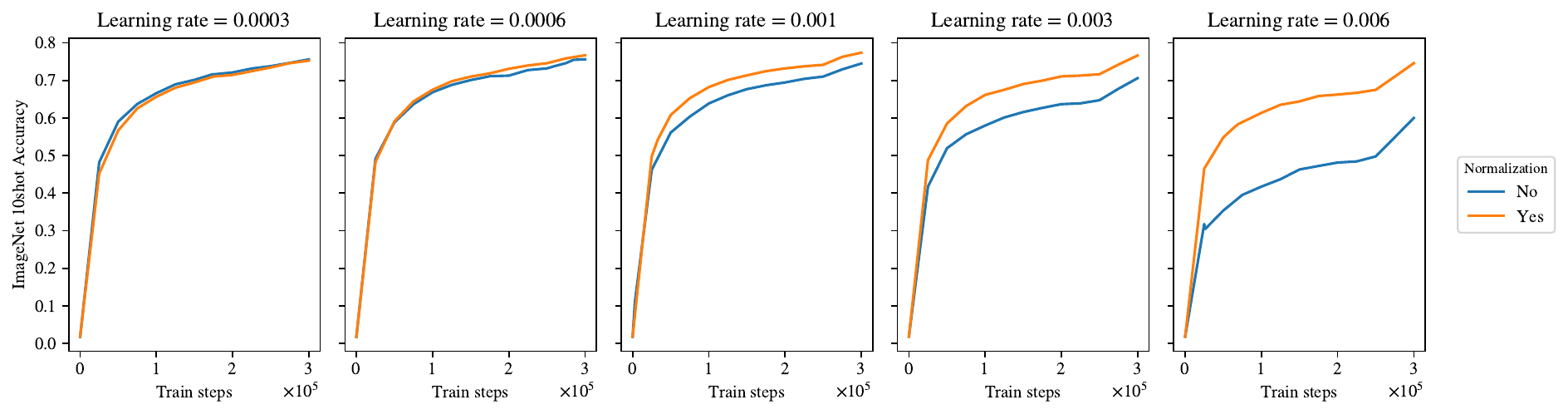}
\caption{ImageNet 10shot accuracy.}
\end{subfigure}
\begin{subfigure}[b]{\textwidth}
\centering
\includegraphics[width=\textwidth]{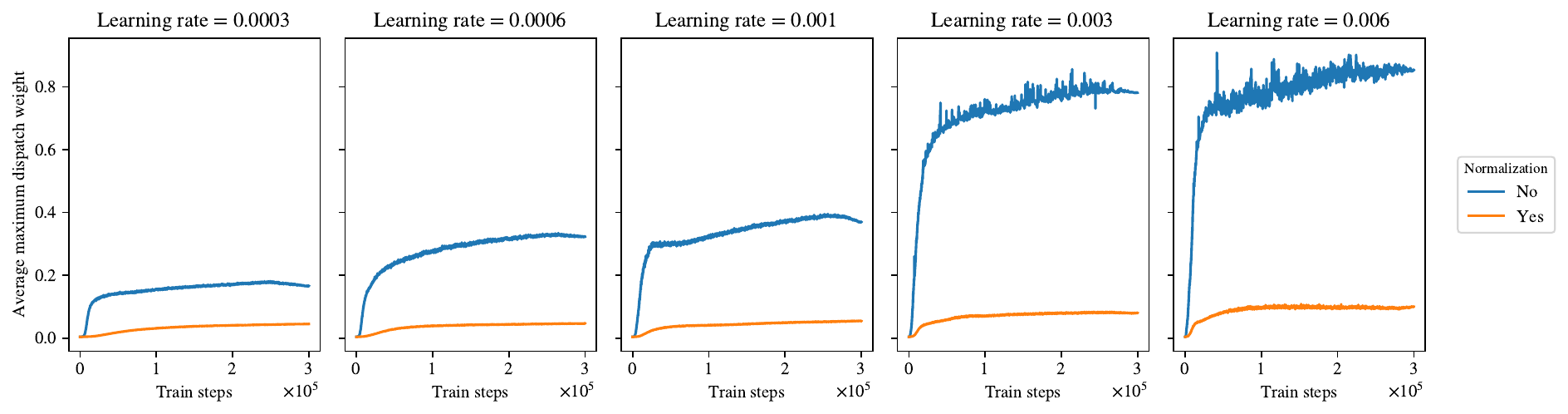}
\caption{Average value of the maximum dispatch weight per slot.}
\end{subfigure}
\begin{subfigure}[b]{\textwidth}
\centering
\includegraphics[width=\textwidth]{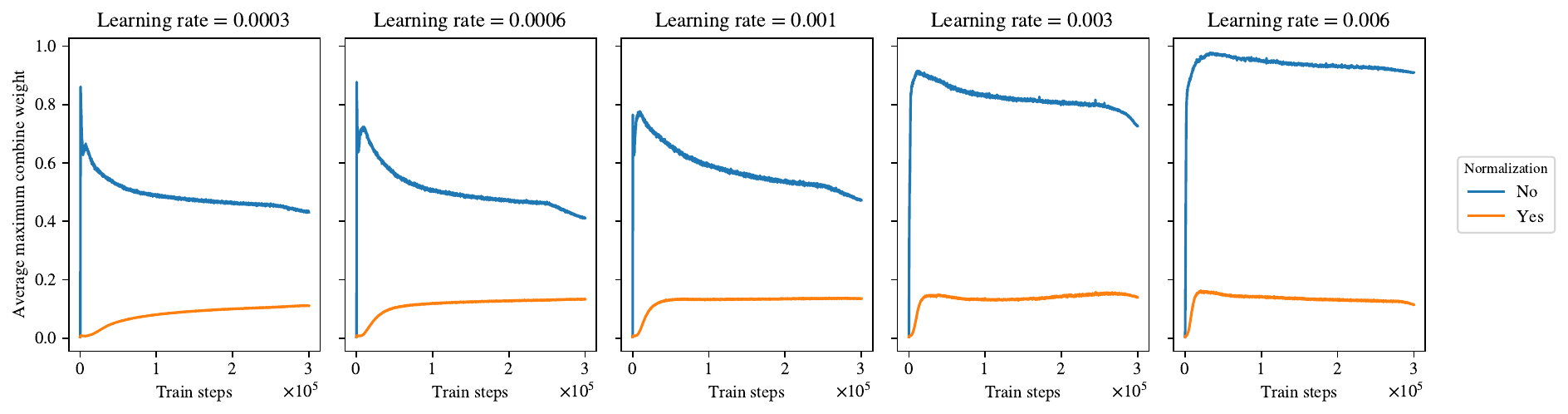}
\caption{Average value of the maximum combine weight per token.}
\end{subfigure}
\caption{%
Training plots of the ImageNet 10shot accuracy (top), the average value of the maximum dispatch weight per slot (middle) and the average value of the maximum combine weight per token (bottom) for different peak values of the learning rate, using a model dimension of $d = 1664$ (i.e. that of a G backbone).
\label{fig:layernorm_issues_learning_rate}}
\end{figure}

In the previous experiment, we trained our model with a linear decay schedule and a peak value of $10^{-3}$. In addition, we also found that applying the softmax layer directly on the output of layer normalization is also very sensible to the learning rate's configuration. Once again, our recipe suggested in \cref{sec:implementation} gives equal or better quality, and is generally more stable.
\Cref{fig:layernorm_issues_learning_rate} shows different metric curves during the training of the same small SoftMoE model as before, with a model dimension of $d = 1664$, using an inverse square root learning rate schedule, with a fixed timescale of $10^5$, a linear warmup phase of $10^5$ steps, and a linear cooldown of $5 \cdot 10^5$ steps, varying the peak learning rate value.
In this figure, similarly to the results from the previous experiment, the average maximum values of the dispatch and combine weights grows to values approaching 1.0 (indicating a collapse in the softmax layers to a one-hot vector), when the inputs to the softmax in the SoftMoE layers are not normalized, which eventually severely hurts the accuracy of the model. However, using the normalization in \cref{sec:implementation} gives better accuracy and makes the model less sensible to the choice of the peak value of the learning rate.

\clearpage
\section{Additional Results}
\label{app:additional_results}

\subsection{Contrastive Learning on LAION-400M}
\label{app:additional_results_laion}
We additionally train a contrastive model, similarly as described on \cref{sec:contrastive_experiments}, but training both the vision and the language towers from scratch on the publicly available dataset LAION-400M\footnote{We use in fact a subset of 275M images and text pairs, since many of the original 400M examples could not be downloaded or contained corrupted images (i.e. the decoding of such images failed).} \citep{schuhmann2021laion}. The backbone architecture of the vision tower is a B/16. We train one model using a plain Vision Transformer, another one with MoE layers on the second half of the network with 128 experts on each layer, using an Experts Choice router, and a third model using \name. All the models use the same text tower architecture without MoEs. All the code necessary to replicate the experiments, including the training hyperparameters used, is available at \url{https://github.com/google-research/vmoe}.

\begin{figure}[htb]
\centering
\includegraphics[width=\textwidth]{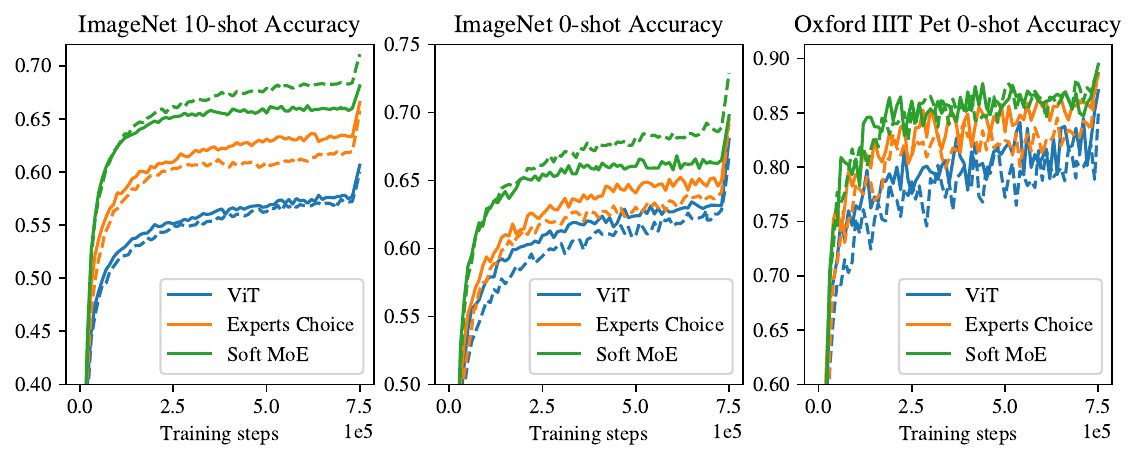}
\caption{\textbf{Results after pretraining on LAION-400M.} Dashed lines correspond to models trained with ``Inception crop'' data augmentation, while solid lines correspond to models trained without any data augmentation. \name performs better than vanilla Vision Transformer (ViT) and Sparse MoE with an Experts Choice router, and benefits from data augmentation.
\label{fig:laion400m_w_wo_inception_crop}}
\end{figure}

\Cref{fig:laion400m_w_wo_inception_crop} shows that, as with JFT-4B and WebLI pre-training (see \cref{sec:classification_experiments,sec:contrastive_experiments} respectively), when pretraining on the publicly available dataset LAION-400M, \name performs significantly better than both vanilla Vision Transformers and ViT with Sparse MoE layers using the Experts Choice router, across different downstream metrics. In addition, we can also see that \name benefits from data augmentation, while neither vanilla ViT or Expert Choice do. Following the observation from \cref{fig:heatmap} in \cref{sec:classification_experiments}, we hypothesize that this is because \name can better utilize the expert parameters.

\subsection{The effect of Batch Priority Routing on Tokens Choice routing}
\Cref{table:bpr_vs_no_bpr} shows the JFT Precision-at-1 and ImageNet 10-shot Accuracy of S/16 MoE models with Tokens Choice router, with/without using Batch Priority Routing (BPR), for different number of total experts and selected experts per token. This table shows that BPR is especially useful for $K = 1$.

\begin{table}[htb]
\begin{center}
\caption{Comparison between Top-K with and without BPR.}
\label{table:bpr_vs_no_bpr}
\begin{tabular}{ cccccc }
\toprule
Model & Number of Experts & K & BPR & JFT prec@1 & IN/10shot \\
\midrule
V-MoE S/16 & 32 & 1 & No & 50.1\% & 64.5\% \\ 
V-MoE S/16 & 32 & 1 & Yes & 51.2\% & 68.9\% \\ 
\midrule
V-MoE S/16 & 32 & 2 & No & 52.5\% & 71.0\% \\ 
V-MoE S/16 & 32 & 2 & Yes & 52.8\% & 71.4\% \\ 
\midrule
V-MoE S/16 & 64 & 1 & No & 50.0\% & 64.4\% \\ 
V-MoE S/16 & 64 & 1 & Yes & 51.5\% & 69.1\% \\ 
\midrule
V-MoE S/16 & 64 & 2 & No & 52.9\% & 70.9\% \\ 
V-MoE S/16 & 64 & 2 & Yes & 52.9\% & 71.4\% \\ 
\bottomrule
\end{tabular}
\end{center}
\end{table}

\subsection{Additional tables and plots complementing Section~\ref{sec:classification_experiments}}

\begin{table}[htb]
\begin{center}
\caption{Training and finetuning results for \name and dense models. Finetuning results on ImageNet at 384 resolution.
We use one slot per expert and did not increase this number during finetuning, thus \names become cheaper than ViT, as the number of input tokens grows to 576 (patch size 16x16) and 752 (patch size 14x14) but the number slots is fixed to a much smaller number (either 128 or 256).}
\label{tab:long_runs_upstream}
\resizebox{\textwidth}{!}{%
\setlength{\tabcolsep}{2pt} %
\begin{tabular}{lrrrrrrrrr}
\toprule
Model & Params & Train steps & Train days & \& exaFLOP & Eval Ms/img & \& GFLOP/img & JFT P@1 & IN/10s & IN/ft\\
\midrule
ViT S/16 & 33M & 4M (50k) & 153.5 & 227.1 & 0.5 & 9.2 & 51.3 & 67.6 & 84.0 \\
\name S/16 128E  & 933M & 4M (50k) & 175.1 & 211.9 & 0.7 & 8.6 & 58.1 & 78.8 & 86.8 \\
\name S/16 128E  & 933M & 10M (50k) & 437.7 & 529.8 & 0.7 & 8.6 & 59.2 & 79.8 & 87.1 \\
\name S/14 256E  & 1.8B & 4M (50k) & 197.9 & 325.7 & 0.9 & 13.2 & 58.9 & 80.0 & 87.2 \\
\name S/14 256E  & 1.8B &  10M (500k) & 494.7 & 814.2 & 0.9 & 13.2 & 60.9 & 80.7 & 87.7 \\
\midrule
ViT B/16  & 108M & 4M (50k) & 410.1 & 864.1 & 1.3 & 35.1 & 56.2 & 76.8 &86.6 \\
\name B/16 128E  & 3.7B & 4M (50k) & 449.5 & 786.4 & 1.5 & 32.0 & 60.0 & 82.0 & 88.0 \\
\midrule
ViT L/16  & 333M & 4M (50k) & 1290.1 & 3025.4 & 4.9 & 122.9 & 59.8 & 81.5 & 88.5 \\
\name L/16 128E  & 13.1B & 1M (50k) & 338.9 & 683.5 & 4.8 & 111.1 & 60.2 & 82.9 & 88.4 \\
\name L/16 128E  & 13.1B & 2M (50k) & 677.7 & 1367.0 & 4.8 & 111.1 & 61.3 & 83.3 & 88.9 \\
\name L/16 128E  & 13.1B & 4M (50k) & 1355.4 & 2734.1 & 4.8 & 111.1 & 61.3 & 83.7 & 88.9 \\
\midrule
ViT H/14  & 669M & 1M (50k) & 1019.9 & 2060.2 & 8.6 & 334.2 & 58.8 & 82.7 & 88.6 \\
ViT H/14  & 669M & 2M (50k) & 2039.8 & 4120.3 & 8.6 & 334.2 & 59.7 & 83.3 & 88.9 \\
\name H/14 128E  & 27.3B & 1M (50k) & 1112.7 & 1754.6 & 8.8 & 284.6 & 61.0 & 83.7 & 88.9 \\
\name H/14 128E  & 27.3B & 2M (50k) & 2225.4 & 3509.2 & 8.8 & 284.6 & 61.7 & 84.2 & 89.1 \\
\name H/14 256E  & 54.1B & 1M (50k) & 1276.9 & 2110.1 & 10.9 & 342.4 & 60.8 & 83.6 & 88.9 \\
\name H/14 256E  & 54.1B & 2M (50k) & 2553.7 & 4220.3 & 10.9 & 342.4 & 62.1 & 84.3 & 89.1 \\
\bottomrule
\end{tabular}}
\end{center}
\end{table}

\begin{figure}[htb]
\centering
\includegraphics[width=\textwidth]{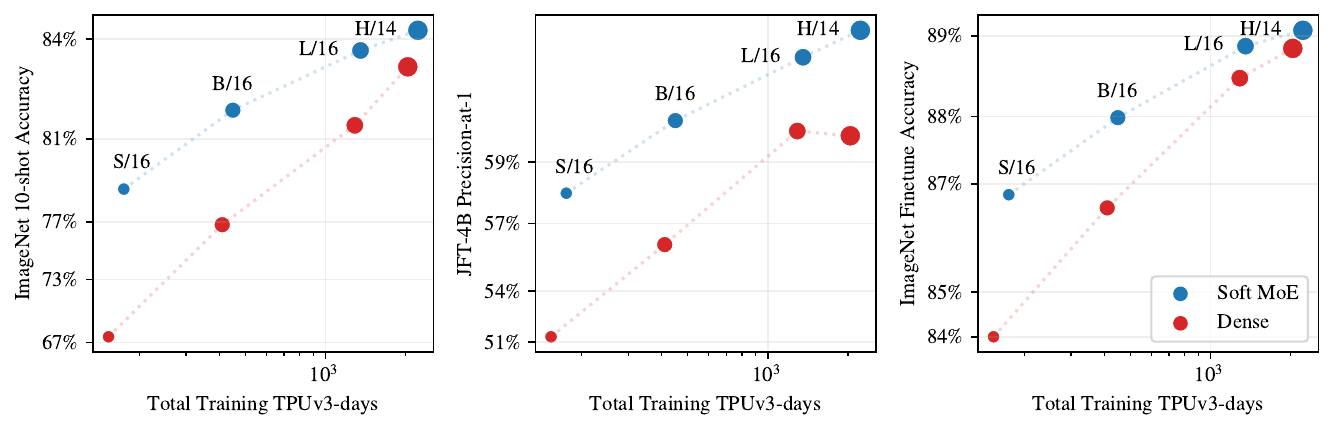}
\caption{%
\textbf{Long runs.}
\name and ViT models trained for 4 million steps with batch size 4096 (H/14 models trained for 2 million steps instead).
Equivalent model classes (S/16, B/16, L/16, H/14) have similar training costs, but \name outperforms ViT on all metrics.
We show ImageNet 10-shot (left), JFT precision at 1 (middle) and ImageNet accuracy after finetuning (right), versus total training FLOPs.
See Table~\ref{tab:long_runs_upstream}.
We report training FLOPs in Figure~\ref{fig:softmoe_vs_vit_long_runs}.
\label{fig:softmoe_vs_vit_long_runs_time}}
\end{figure}

\begin{figure}[htb]
\centering
\includegraphics[width=\textwidth]{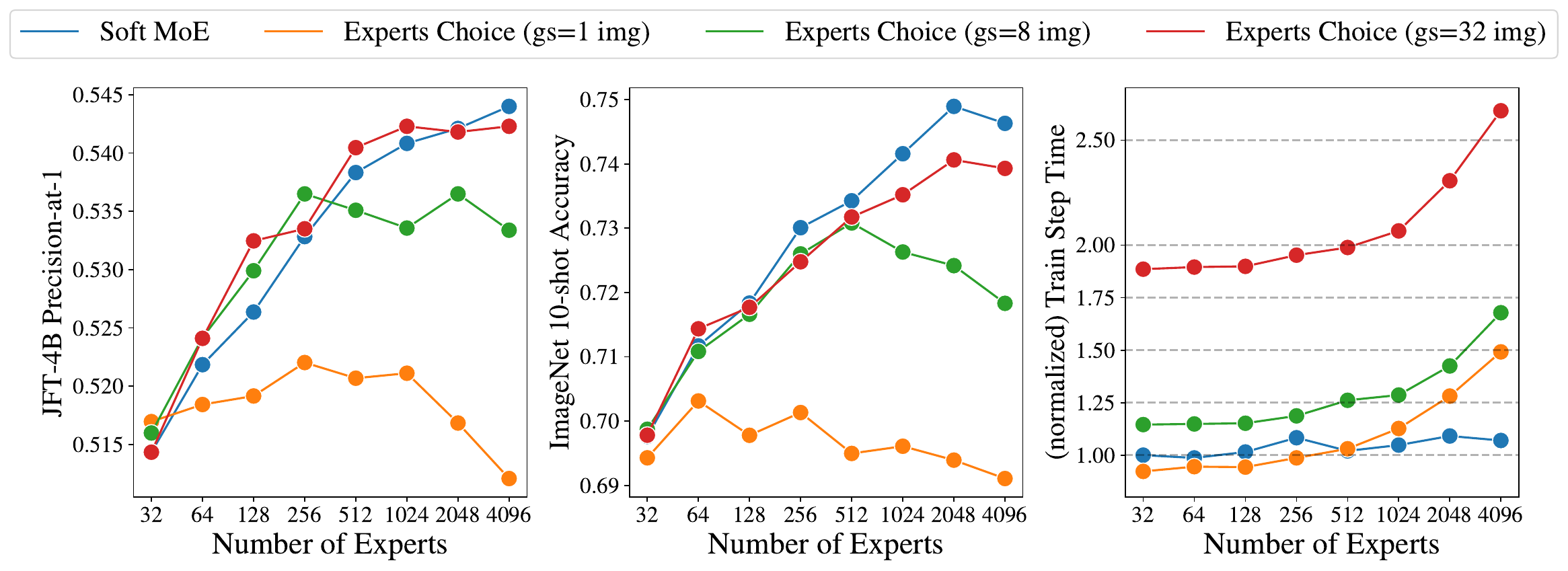}
\caption{JFT precision-at-1, ImageNet 10-shot accuracy, and normalized training step time when increasing the total number of experts while keeping the total amount of slots fixed. \name achieves consistently better results with more experts, whereas cost is kept roughly constant. Adding too many experts to Experts Choice hurt performance and significantly increases the cost.
Experts Choice can perform well with many experts if we increase the group size up to 32 images per group.
The normalized train step time is computed with respect to \name with 32 experts.
Experts Choice with 32 images per group and 4096 experts requires more than 2.5x its cost.
\label{fig:ablation_total_slots_top_c_gs}}
\end{figure}

\begin{figure}[htb]
\centering
\includegraphics[width=\textwidth]{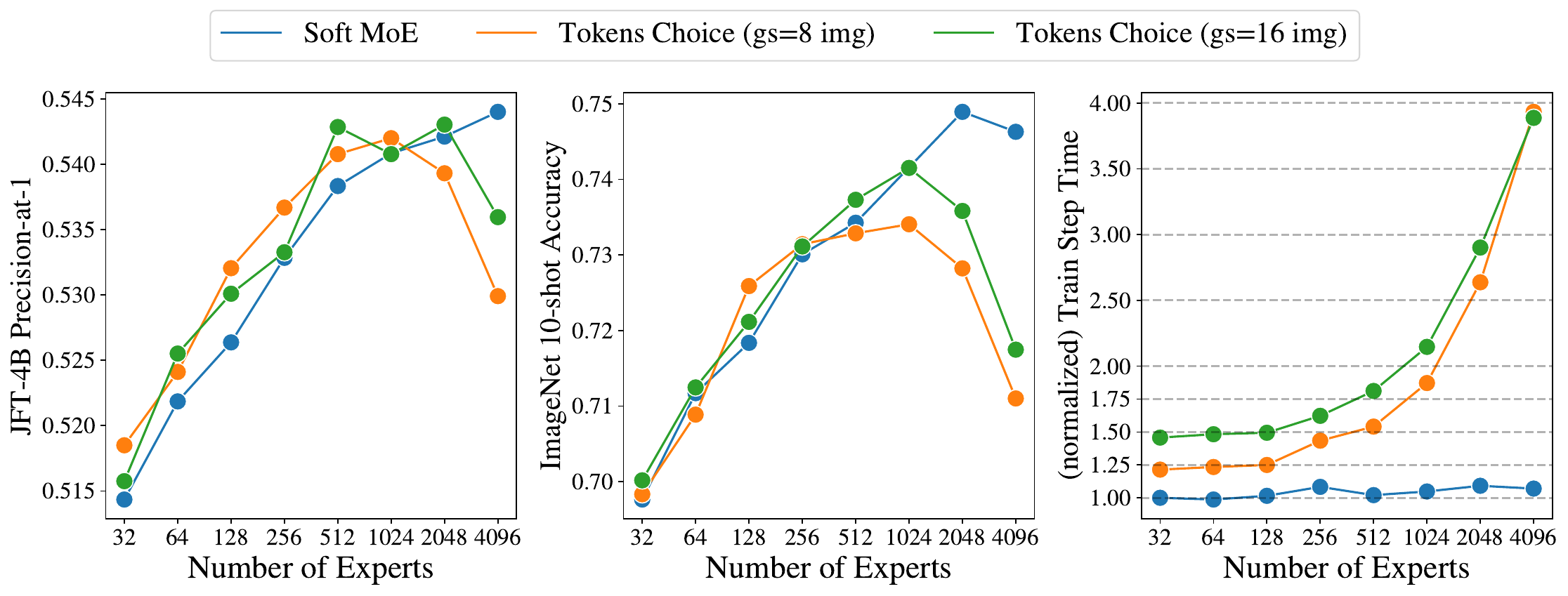}
\caption{JFT precision-at-1, ImageNet 10-shot accuracy, and normalized training step time when increasing the total number of experts while keeping the total amount of slots fixed. \name achieves consistently better results with more experts, whereas cost is kept roughly constant. Adding too many experts to Tokens Choice hurt performance and significantly increases the cost.
Even with a large group size (16 images), Tokens Choice struggles to perform well with a few thousand experts.
The normalized train step time is computed with respect to \name with 32 experts.
Tokens Choice with 8 or 16 images per group and 4096 experts requires almost 4x its cost.
\label{fig:ablation_total_slots_top_k_gs}}
\end{figure}

\begin{figure}[htb]
\centering
\begin{subfigure}[b]{0.49\textwidth}
\includegraphics[width=\textwidth]{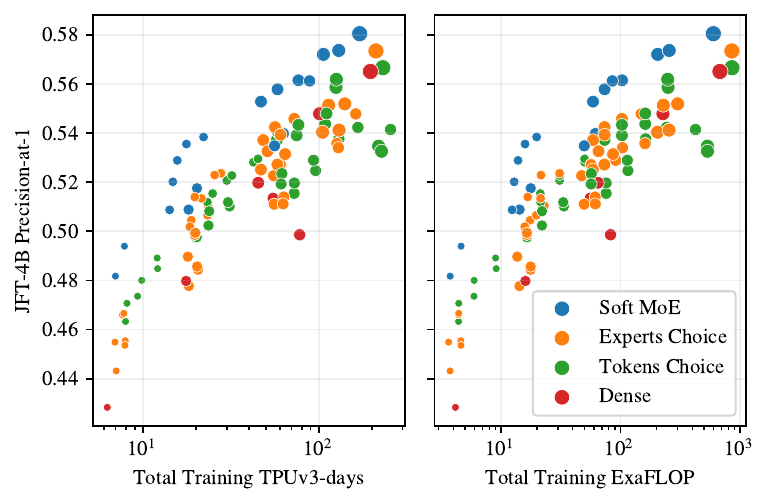}
\caption{JFT-4B Precision-at-1\label{fig:upstream_pareto_dom}}
\end{subfigure}
\hfill
\begin{subfigure}[b]{0.49\textwidth}
\includegraphics[width=\textwidth]{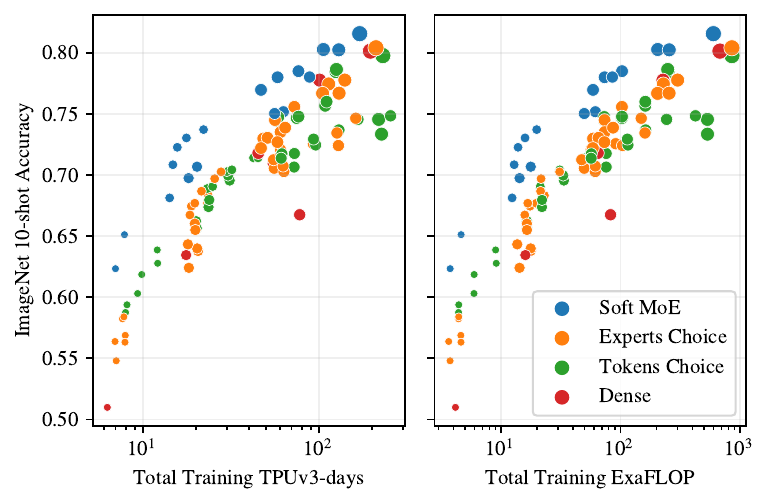}
\caption{ImageNet 10-shot Accuracy\label{fig:imagenet10shot_pareto_dom}}
\end{subfigure}
\caption{JFT-4B Precision-at-1 and ImageNet 10-shot accuracy on short runs (300k steps). The size of the marker depends on the backbone size: S/32, S/16, B/32, B/16, L/16 and H/14. Colors represent different methods: \name (blue), Sparse MoEs with Experts Choice (orange) and Tokens Choice routing (green), and a Dense (red) model. MoE runs include different configurations.
\label{fig:small_pareto_all}}
\end{figure}

\begin{figure}[htb]
\centering
\begin{subfigure}[b]{0.49\textwidth}
\includegraphics[width=\textwidth]{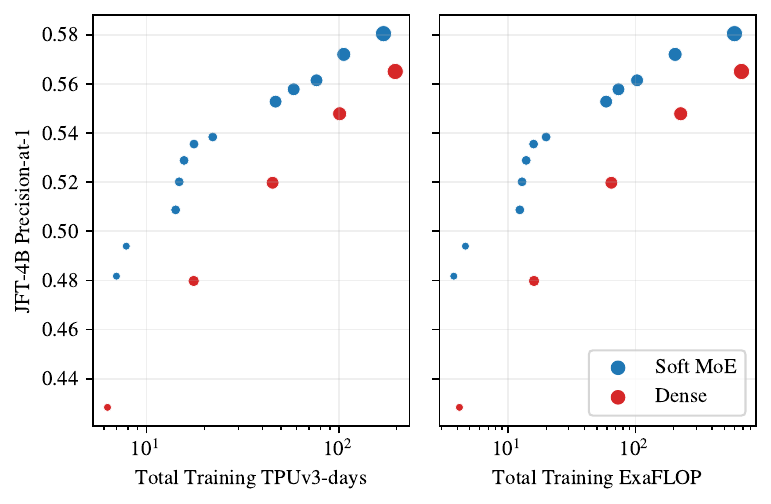}
\caption{JFT-4B Precision-at-1\label{fig:upstream_soft_dense}}
\end{subfigure}
\hfill
\begin{subfigure}[b]{0.49\textwidth}
\includegraphics[width=\textwidth]{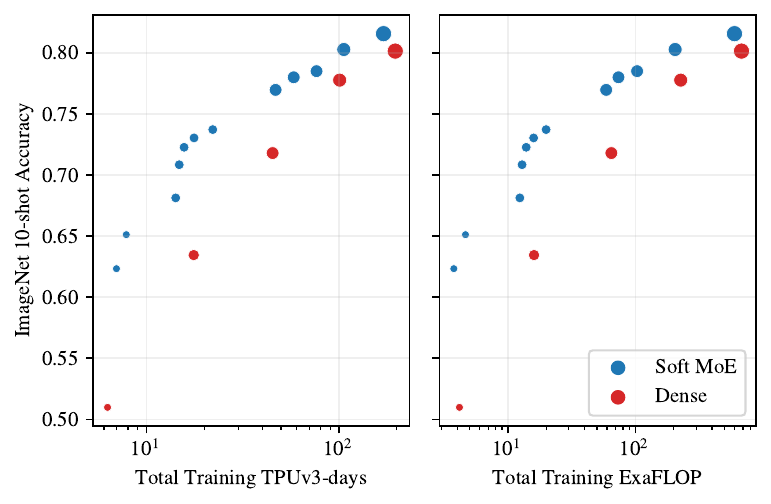}
\caption{ImageNet 10-shot Accuracy\label{fig:imagenet10shot_soft_dense}}
\end{subfigure}
\caption{JFT-4B Precision-at-1 and ImageNet 10-shot accuracy on short runs (300k training steps). The size of the marker depends on the backbone size: S/32, S/16, B/32, B/16, L/16 and H/14. Colors represent different methods: \name (blue) and Dense (red) models. MoE runs include different configurations. We only show the runs that are not dominated by another model using the same method (S/8 and L/32 were always dominated).
\label{fig:small_soft_dense}}
\end{figure}

\begin{figure}[htb]
\centering
\begin{subfigure}[b]{0.49\textwidth}
\includegraphics[width=\textwidth]{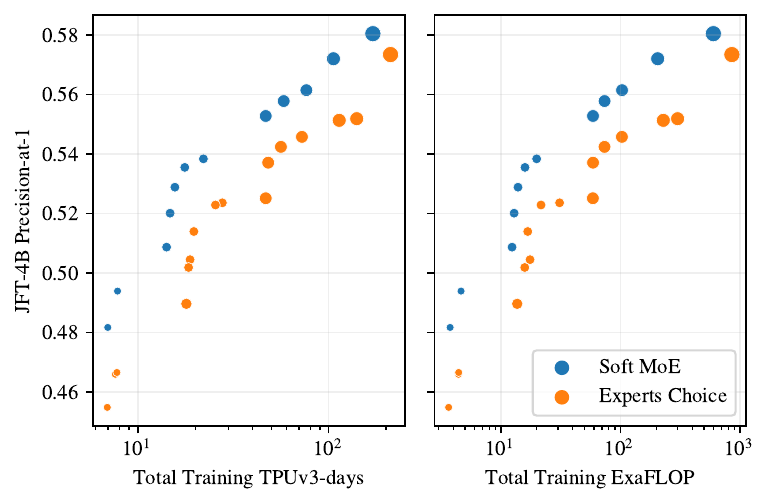}
\caption{JFT-4B Precision-at-1\label{fig:upstream_soft_topc}}
\end{subfigure}
\hfill
\begin{subfigure}[b]{0.49\textwidth}
\includegraphics[width=\textwidth]{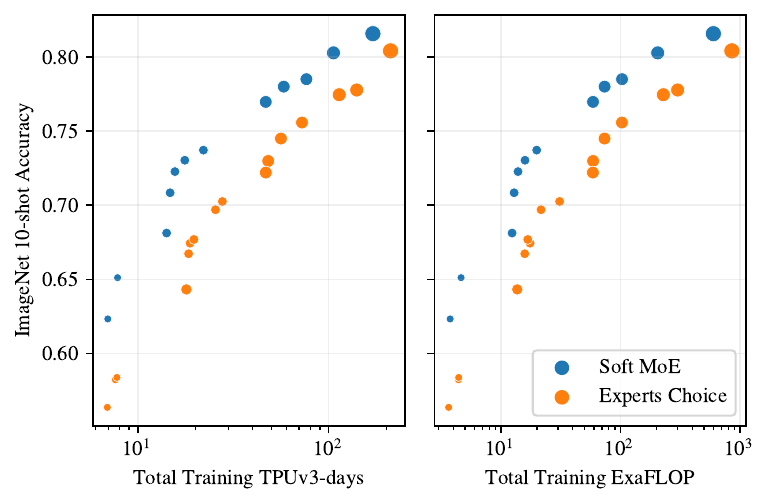}
\caption{ImageNet 10-shot Accuracy\label{fig:imagenet10shot_soft_topc}}
\end{subfigure}
\caption{JFT-4B Precision-at-1 and ImageNet 10-shot accuracy on short runs (300k training steps). The size of the marker depends on the backbone size: S/32, S/16, B/32, B/16, L/16 and H/14. Colors represent different methods: \name (blue) and Sparse MoEs with Experts Choice (orange) models. MoE runs include different configurations. We only show the runs that are not dominated by another model using the same method (S/8 and L/32 were always dominated).
\label{fig:small_soft_topc}}
\end{figure}

\begin{figure}[htb]
\centering
\begin{subfigure}[b]{0.49\textwidth}
\includegraphics[width=\textwidth]{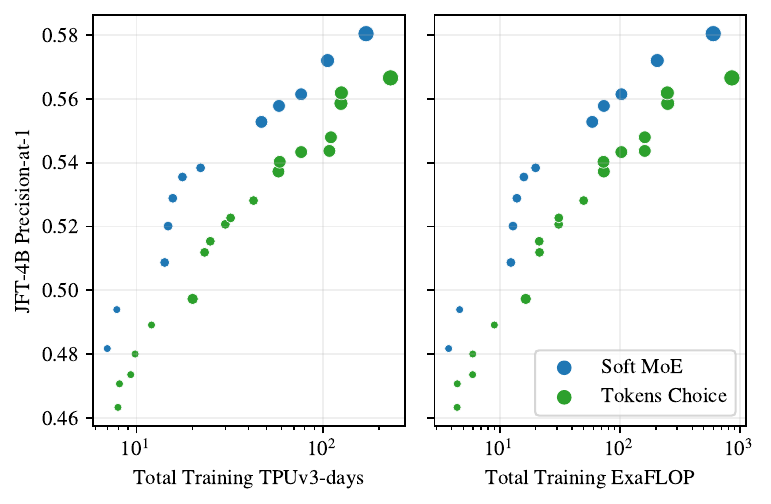}
\caption{JFT-4B Precision-at-1\label{fig:upstream_soft_topk}}
\end{subfigure}
\hfill
\begin{subfigure}[b]{0.49\textwidth}
\includegraphics[width=\textwidth]{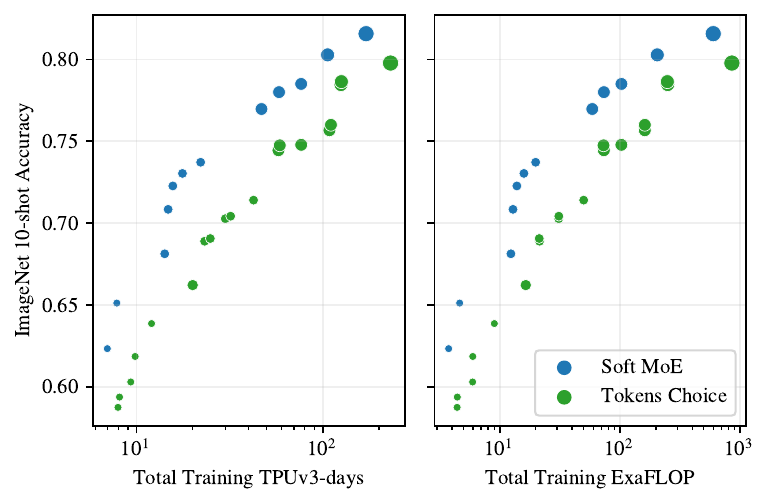}
\caption{ImageNet 10-shot Accuracy\label{fig:imagenet10shot_soft_topk}}
\end{subfigure}
\caption{JFT-4B Precision-at-1 and ImageNet 10-shot accuracy on short runs (300k training steps). The size of the marker depends on the backbone size: S/32, S/16, B/32, B/16, L/16 and H/14. Colors represent different methods: \name (blue) and Sparse MoEs with Tokens Choice (green) models. MoE runs include different configurations. We only show the runs that are not dominated by another model using the same method (S/8 and L/32 were always dominated).
\label{fig:small_soft_topk}}
\end{figure}

\begin{figure}[htb]
\centering
\includegraphics[width=\textwidth]{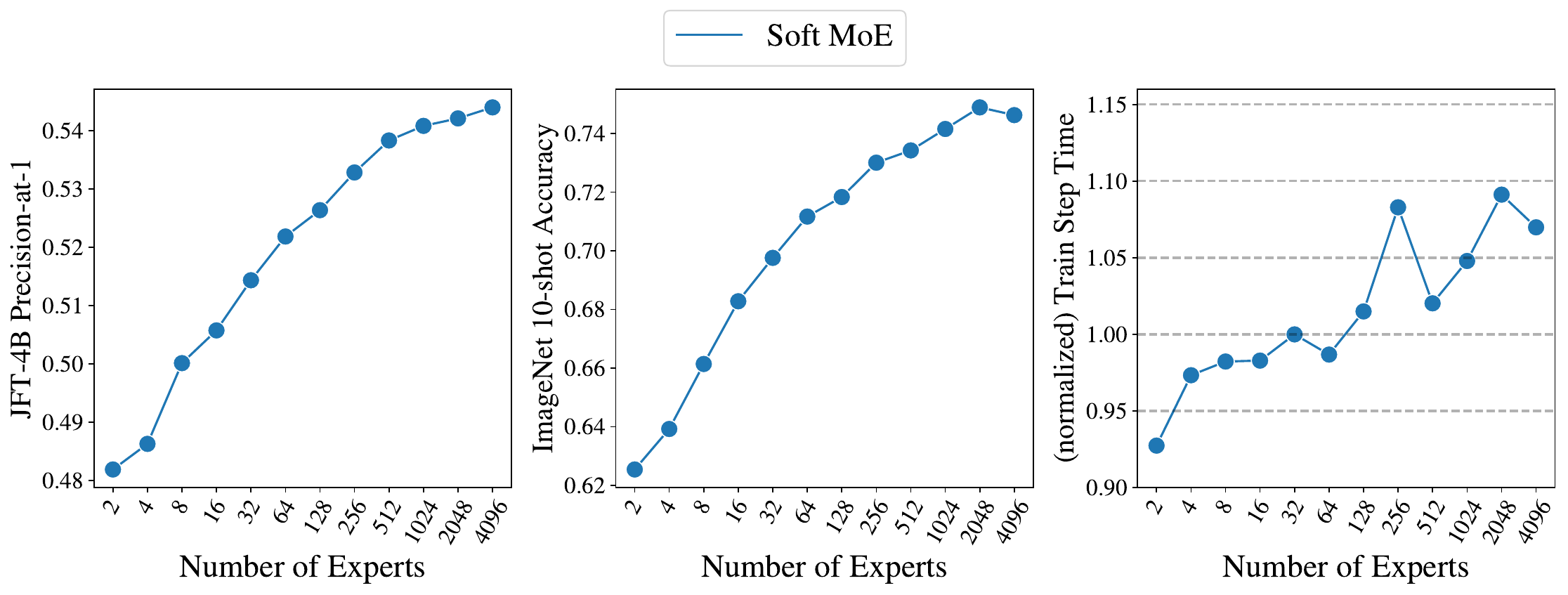}
\caption{\textbf{JFT Precision-at-1, ImageNet 10-shot Accuracy, and normalized Training Step time when increasing the total number of experts while keeping the total amount of slots fixed (4096)}. \name achieves consistently better results with more experts, whereas cost is kept roughly constant (same FLOPs but communication costs vary due to higher topologies needed for larger models).
The normalized train step time is computed with respect to \name with 32 experts.
Model sizes range from 38M (2 experts) to 9.7B parameters (4096 experts).
\label{fig:ablation_total_slots_softmoe}}
\end{figure}

\clearpage
\section{Model Inspection}
\label{app:model_inspection}
In this section, we take a look at various aspects of the routing the model learns.

\begin{figure*}[htb]
\centering
\includegraphics[width=\textwidth]{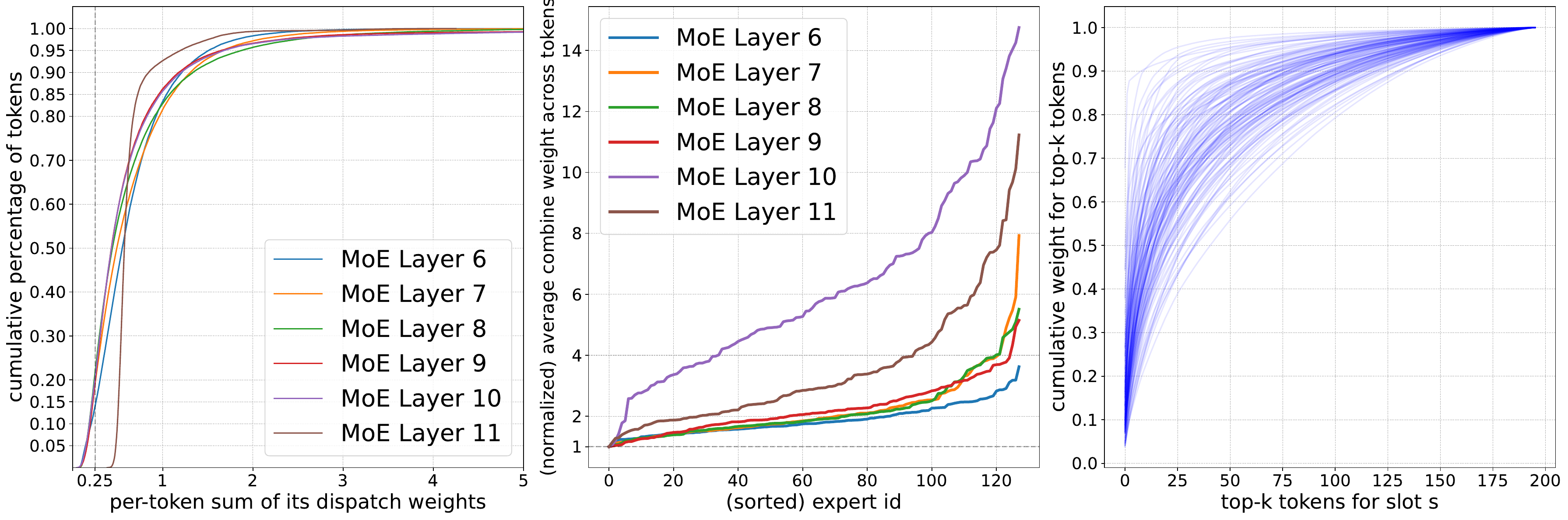}
\caption{\textbf{(Left)} Distribution of summed dispatch weights per token for different MoE layers. For instance, in layer 11, the dispatch weights for 90-95\% of the input tokens summed over all the slots are at most 1. Only a tiny fraction of tokens contribute to slots by summing more than 3. \textbf{(Middle)} Distribution of combine weights per slot (or expert, as we use one slot per expert) summed across all input tokens. We normalize the sum by its minimum value across experts. \textbf{(Right)} Each curve corresponds to one slot. Dispatch weights from all tokens to each slot add up to 1. Distribution of how many inputs tokens are needed to achieve a certain fraction of the total weight for the slot.}
\label{fig:model_inspection_stats}
\end{figure*}

\textbf{Tokens contributions to slots.}
While there is no dropping in \name, it is still possible that some tokens contribute little to \emph{all} slots if their logits are much lower than those of other tokens.
We would like to see if some tokens contribute to slots in a disproportionate manner.
\Cref{fig:model_inspection_stats} (left) shows the distribution across tokens for the total weight each token provides to slots (i.e.\ summed over all slots).
This was computed over a batch with 256 images with 196 tokens each on a \name S/16 finetuned on ImageNet.
We see there is a heavy tail of tokens that provide a stronger total contribution to slots, and the shape is somewhat similar across layers.
Around 2-5\% of the tokens provide a summed weight above 2.
Also, between 15\% and 20\% of the tokens only contribute up to 0.25 in total weight.
The last layer is slightly different, where token contribution is softer tailed.
\Cref{sec:app_analysis} further explores this.

\textbf{Experts contributions to outputs.}
Similarly, we would like to understand how much different slots end up contributing to the output tokens.
We focus on the case of one slot per expert.
We can approximate the total contribution of each expert (equivalently, slot) by averaging their corresponding coefficients in the linear combinations for all output tokens in a batch.
\Cref{fig:model_inspection_stats} (center) shows such (normalized) importance across experts for different MoE layers.
We see that, depending on the layer, some experts can impact output tokens between 3x and 14x more than others.

\textbf{Number of input tokens per slot.}
For each slot, \cref{fig:model_inspection_stats} (right) shows how many input tokens are required to achieve a certain cumulative weight in its linear combination.
The distribution varies significantly across slots.
For a few slots the top 20-25 tokens account for 90\% of the slot weight, while for other slots the distribution is more uniform and many tokens contribute to fill in the slot.
In general, we see that slots tend to mix a large number of tokens unlike in standard Sparse MoEs.

\begin{figure}[htb]
\centering
\includegraphics[width=\textwidth]{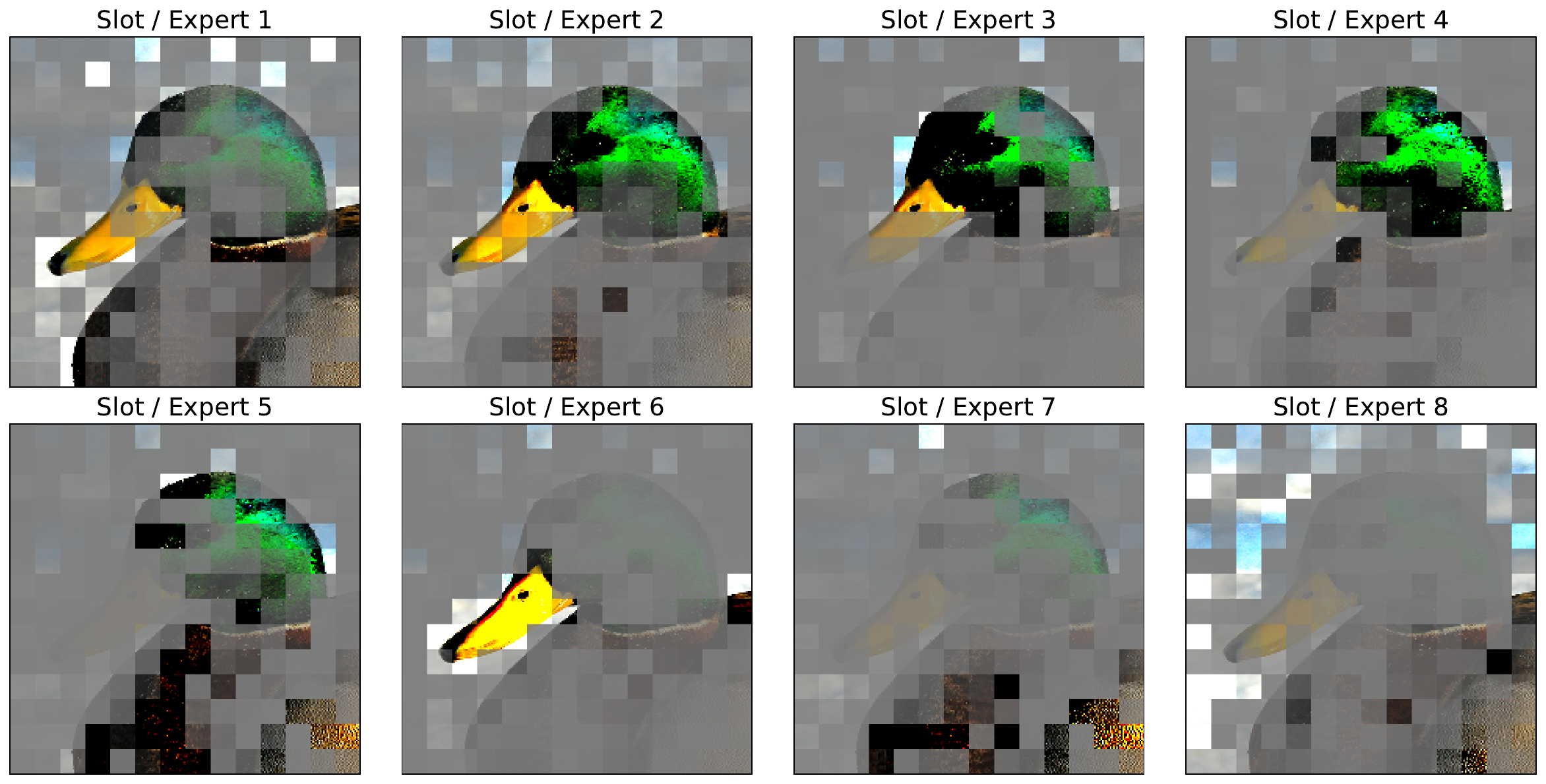}
\caption{Linear combinations for 8 slots when using input image in \cref{fig:sparse_vs_soft_diagram}. Model is \name S/16 with 128 experts and one slot per expert, and it was finetuned on ImageNet. We show results for the first MoE layer (seventh block). The selected slots (among 128) are cherry-picked to highlight differences across slots.}
\label{fig:model_inspection_slots}
\end{figure}

\textbf{Visual inspection.}
In order to provide some intuition regarding how slots average input tokens, \cref{fig:model_inspection_slots} graphically shows the linear combinations for 8 different slots for the image shown in \cref{fig:sparse_vs_soft_diagram}.
We shade patches inversely proportionally to their weight in the slots; note that all tokens representations are eventually combined into a single one (with hidden dimension $h$) before being passed to the expert (unlike in our plot, where they are arranged in the usual way).
These plots correspond to a \name S/16 with 128 experts and one slot per expert, and we handpicked 8 out of the 128 slots to highlight how different slots tend to focus on different elements of the image.

\section{Additional analysis}
\label{sec:app_analysis}

\subsection{Cumulative sum of dispatch and combine weights}

\begin{figure}[tb]
\centering
\includegraphics[width=\textwidth]{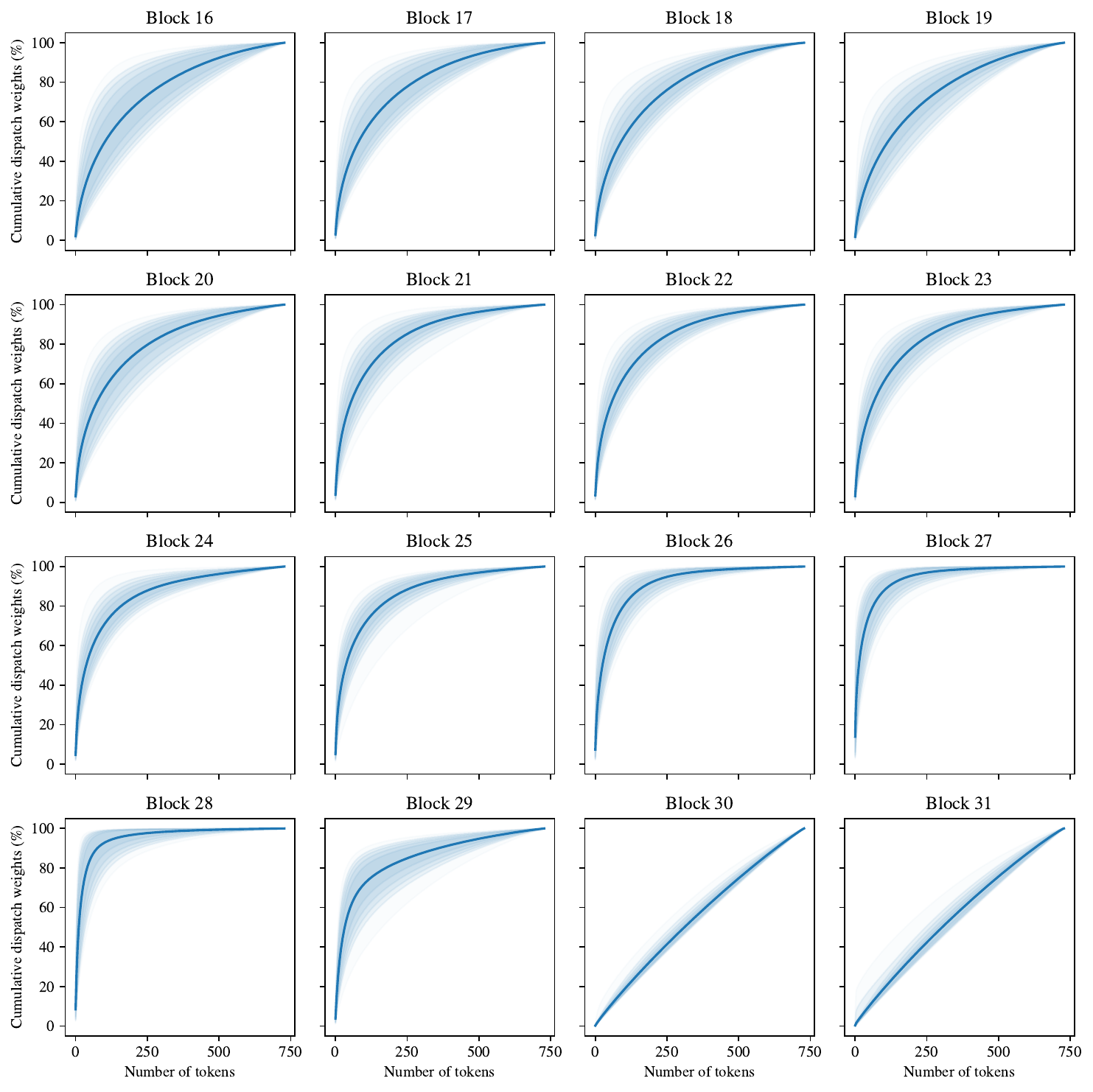}
\caption{\textbf{Distribution of the cumulative sum of dispatch weights.} For each input slot, we compute the cumulative sum of its corresponding dispatch weights (sorted by decreasing value). This indicates over how many input tokens a certain cumulative weight is distributed over.
The line in each plot represents the average computed over all slots and ImageNet validation images of the given block in the SoftMoE H/14 model. The colored areas represent the central 60\%, 80\%, 90\%, 95\% and 99\% of the distribution (from darker to lighter, better seen in color).
\label{fig:analysis_h14_cumulative_dispatch_weights}}
\end{figure}

\Cref{fig:analysis_h14_cumulative_dispatch_weights} shows the distribution over slots of the cumulative sum (over tokens) of their corresponding dispatch weights. For each slot we compute the cumulative sum of the dispatch weights over tokens sorted in decreasing order. This indicates how many tokens are necessary to cover a given percentage of the total mass of the weighted average. We compute this cumulative sum for all slots over all the 50\,000 ImageNet validation images, across all layers of the \name H/16 model after finetuning. In the plot, we represent with a solid line the average (over all slots and images) cumulative sum, and the different colored areas represent the central 60\%, 80\%, 90\%, 95\% and 99\% of the distribution (from darker to lighter colors) of cumulative sums.

This tells us, for instance, how uniform is the weighted average over tokens used to compute each input slot. In particular, each slot in the last two layers is close to a uniform average of all the tokens (a completely uniform average would be represented by a straight line). This tells us that in these layers, every expert processes roughly the same inputs, at least after the model is trained.
However, this weighted average is far from uniform in the rest of the layers, meaning that there are tokens that contribute far more than others. For example, in layer 28, a few tens of tokens already cover 80\% of the weighted average mass.
Finally, given the width of the colored areas, we can also see that there's a significant difference on the weighted averages depending on the slot, across all layers (except maybe the last two). This indicates that the dispatch weights vary across different slots and images.

\begin{figure}[tb]
\centering
\includegraphics[width=\textwidth]{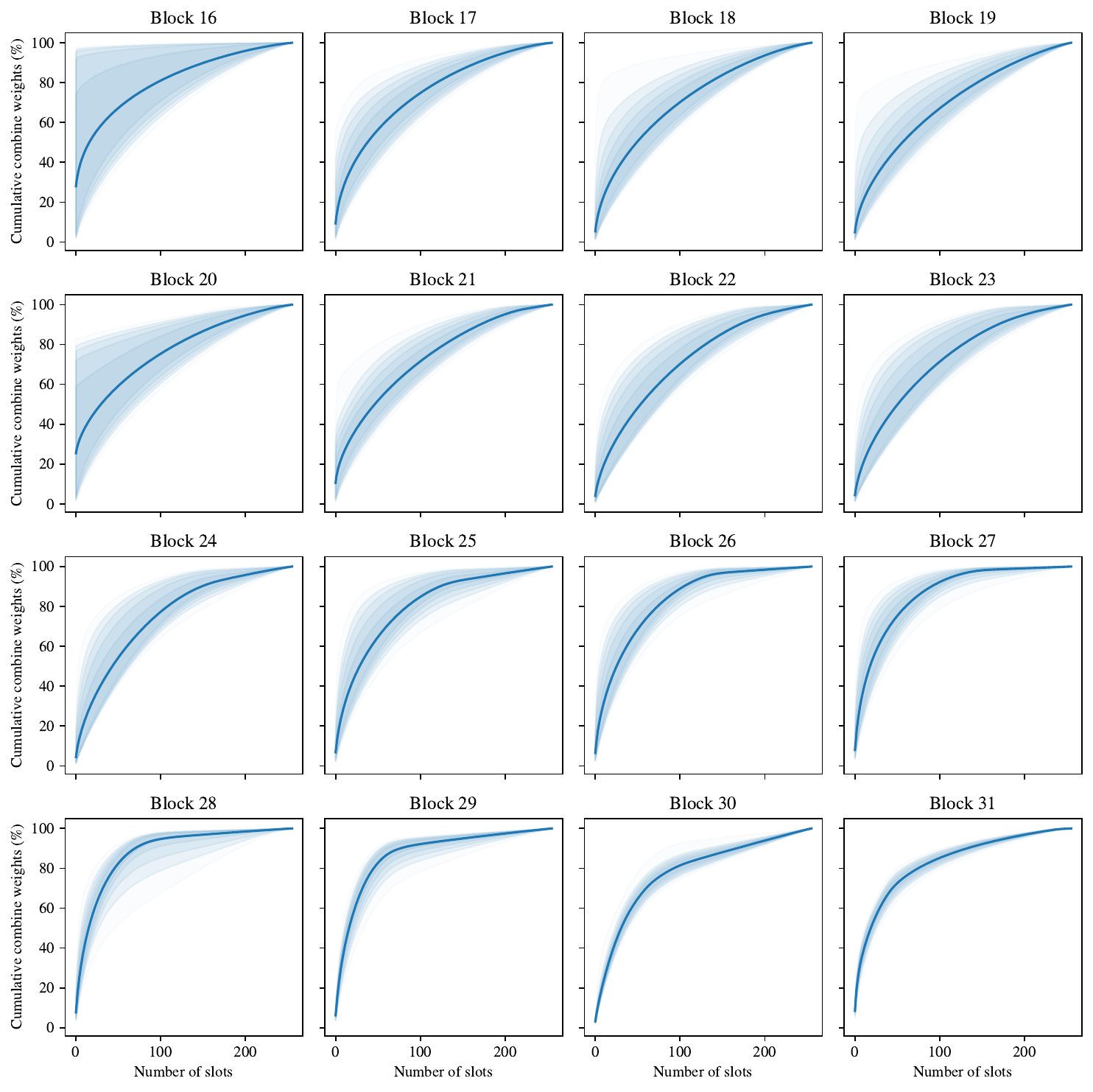}
\caption{\textbf{Distribution of the cumulative sum of combine weights.} For each output token, we compute the cumulative sum of its corresponding combine weights (sorted by decreasing value). This indicates over how many output slots a certain cumulative weight is distributed over.
The line in each plot represents the average computed over all tokens and ImageNet validation images of the given block in the SoftMoE H/14 model. The colored areas represent the central 60\%, 80\%, 90\%, 95\% and 99\% of the distribution (from darker to lighter, better seen in color).
\label{fig:analysis_h14_cumulative_combine_weights}}
\end{figure}

Similarly, \cref{fig:analysis_h14_cumulative_combine_weights} shows the corresponding plots for the cumulative sum of the combine weights. In this case, for each output token we compute the cumulative sum of the combine weights over slots sorted in decreasing order. 
Notice that, although the dispatch weights in the last two layers were almost uniform, the combine weights are not. This indicates that some slots (and thus, experts) are more important than others in computing the output tokens, and thus their corresponding expert parameters are not redundant. Of course, the identity of the ``important'' slots may vary depending on the input token.

\section{Slot Correlation}
\label{app:slot_correlation}
\begin{figure}[htb]
\centering
\includegraphics[width=\textwidth]{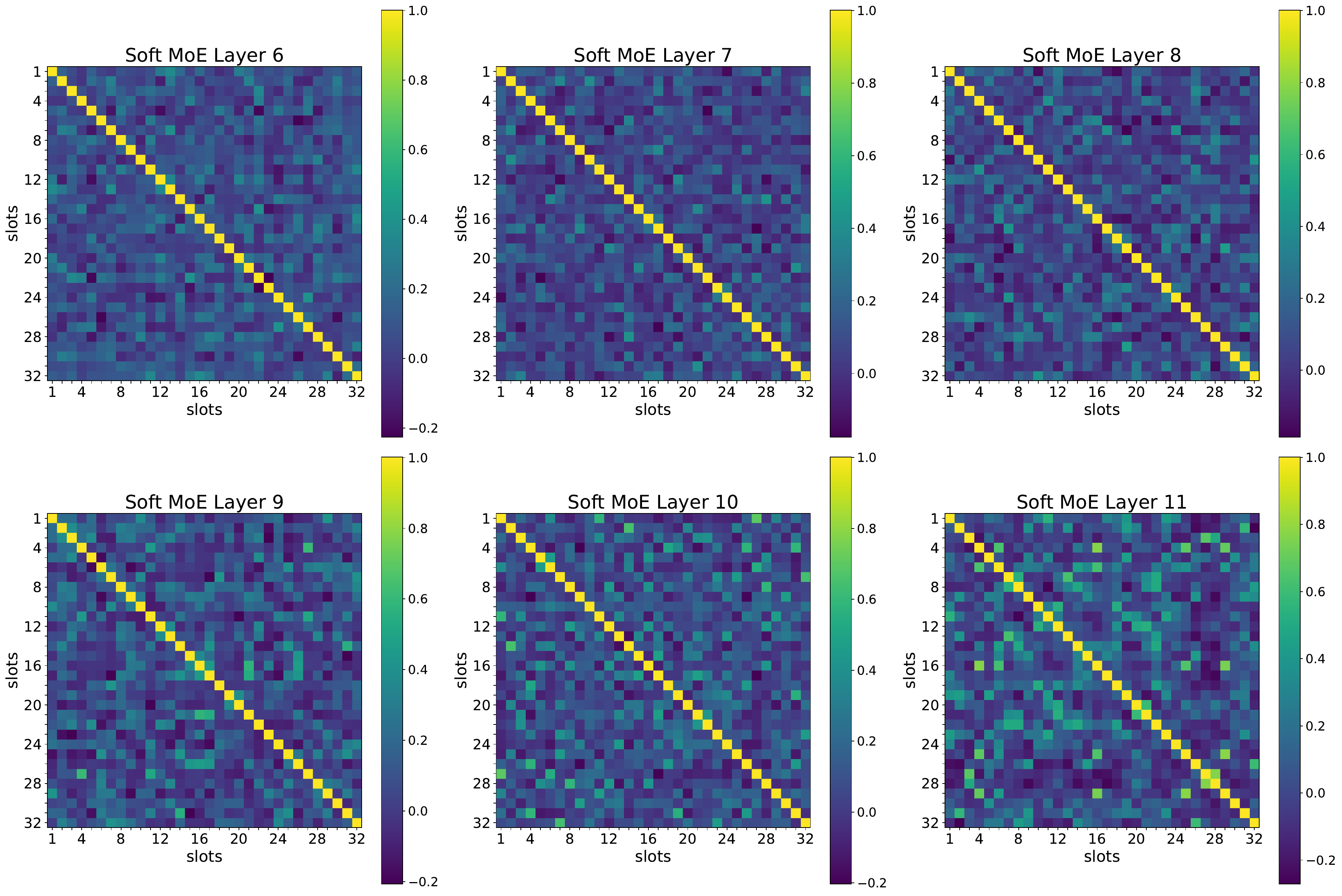}
\caption{{\name S/16 with 1 slot per expert.}
\label{fig:slot_correlation_1s}}
\end{figure}

\begin{figure}[tb]
\centering
\includegraphics[width=\textwidth]{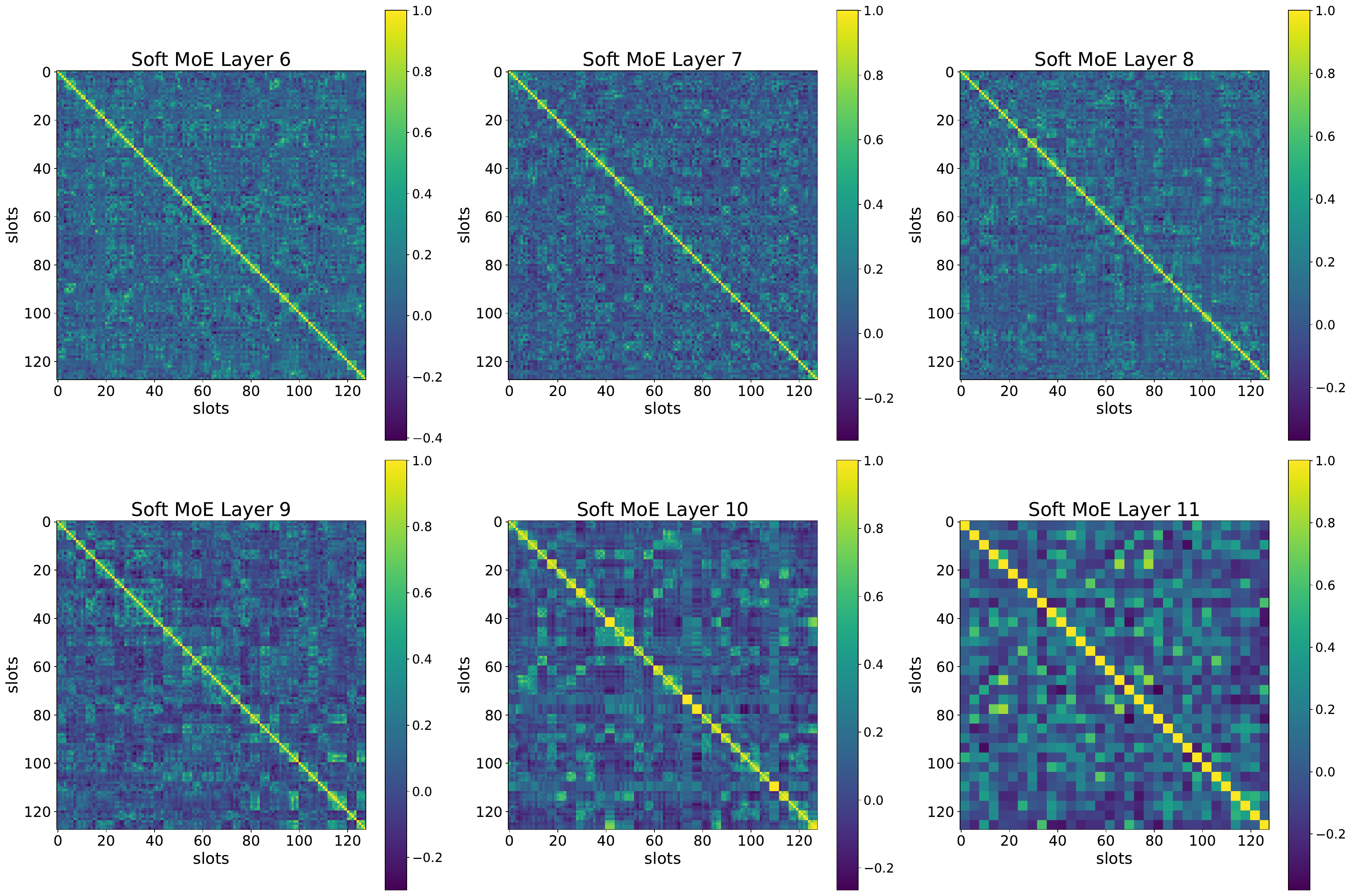}
\caption{{\name S/16 with 4 slots per expert.}
\label{fig:slot_correlation_4s}}
\end{figure}

\begin{figure}[tb]
\centering
\includegraphics[width=\textwidth]{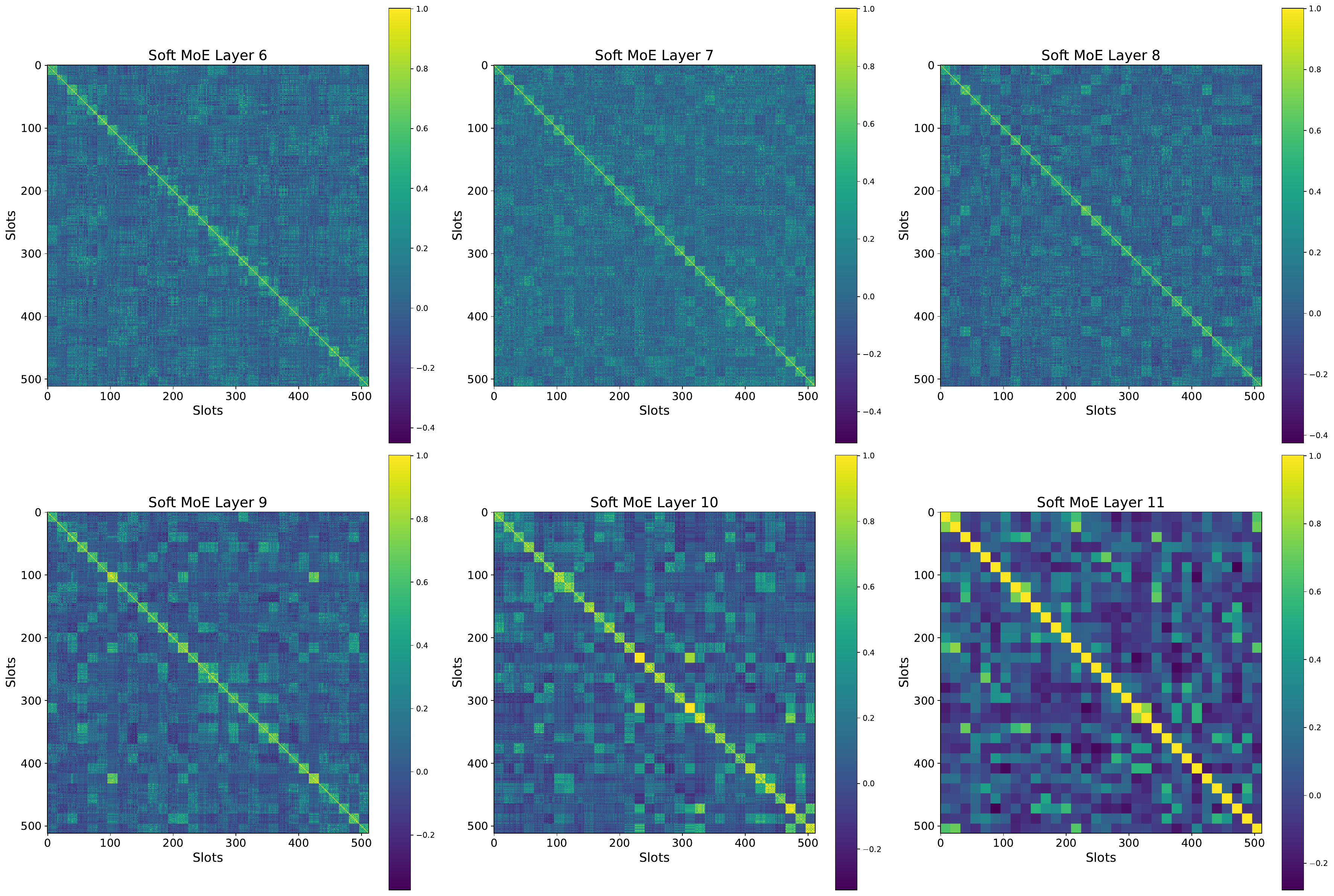}
\caption{{\name S/16 with 16 slots per expert.}
\label{fig:slot_correlation_16s}}
\end{figure}

In this section we explore the correlation between the different slot \emph{parameters} that \name learns, and its relationship with the number of slots per expert.
\Cref{fig:slot_correlation_1s,fig:slot_correlation_4s,fig:slot_correlation_16s} show for each of 6 layers in a \name S/16 the inner product between each pair of (normalized) slot parameter vectors.

While \cref{fig:slot_correlation_1s} shows no clear relationship between slots from different experts (as each expert only has one slot), we observe in \cref{fig:slot_correlation_4s,fig:slot_correlation_16s} how consecutive slots (corresponding to the same expert) are extremely aligned.
This confirms our hypothesis that adding more slots to experts does not work very well as these slots end up aligning their value, and computing somewhat similar linear combinations. Therefore, these projections do not add too much useful information to the different tokens to be processed by the experts (in the extreme, these slots would be identical).

\clearpage
\section{Pareto Models}
\label{sec:app_pareto_runs}

\setlength{\tabcolsep}{2pt} %
\begin{longtable}[c]{lllrrrrrrrr}
\caption{%
\looseness=-1
Model runs from \cref{sec:pareto_plots} (shown in Pareto plot) trained for 300k steps on JFT with inverse square root decay and 50k steps cooldown.
We trained dense and MoE (\name, Tokens Choice, Experts Choice) models with sizes S/32, S/16, S/8, B/32, B/16, L/32, L/16 and H/14. Sorted by increasing training TPUv3 days.}
\label{table:pareto_runs}\\
\toprule
Ref & Model & Routing & Experts & Group Size & K & C & JFT P@1 & IN/10shot & Train exaFLOP & Train Days \\
\midrule \endhead 
\bottomrule  \endfoot
1 & S/32 & Dense & -- & -- & -- & -- & 42.8 & 51.0 & 4.2 & 6.3 \\
2 & S/32 & Experts Choice & 32 & 392 & -- & 0.5 & 45.5 & 56.4 & 3.7 & 6.9 \\
3 & S/32 & Soft MoE & 32 & 49 & -- & -- & 48.2 & 62.3 & 3.8 & 7.0 \\
4 & S/32 & Experts Choice & 32 & 49 & -- & 0.5 & 44.3 & 54.8 & 3.8 & 7.0 \\
5 & S/32 & Experts Choice & 32 & 392 & -- & 1.0 & 46.6 & 58.2 & 4.4 & 7.6 \\
6 & S/32 & Experts Choice & 64 & 392 & -- & 1.0 & 46.7 & 58.4 & 4.4 & 7.8 \\
7 & S/32 & Soft MoE & 64 & 49 & -- & -- & 49.4 & 65.1 & 4.6 & 7.8 \\
8 & S/32 & Experts Choice & 64 & 49 & -- & 1.0 & 45.4 & 56.3 & 4.6 & 7.9 \\
9 & S/32 & Experts Choice & 32 & 49 & -- & 1.0 & 45.5 & 56.9 & 4.6 & 7.9 \\
10 & S/32 & Tokens Choice & 32 & 392 & 1 & 1.0 & 46.3 & 58.7 & 4.4 & 8.0 \\
11 & S/32 & Tokens Choice & 64 & 392 & 1 & 1.0 & 47.1 & 59.4 & 4.4 & 8.1 \\
12 & S/32 & Tokens Choice & 32 & 392 & 2 & 1.0 & 47.4 & 60.3 & 6.0 & 9.3 \\
13 & S/32 & Tokens Choice & 64 & 392 & 2 & 1.0 & 48.0 & 61.9 & 6.0 & 9.8 \\
14 & B/32 & Dense & -- & -- & -- & -- & 48.0 & 63.4 & 16.0 & 11.7 \\
15 & B/32 & Soft MoE & 32 & 49 & -- & -- & 50.9 & 69.7 & 14.3 & 11.7 \\
16 & S/32 & Tokens Choice & 64 & 392 & 2 & 2.0 & 48.9 & 63.9 & 9.0 & 12.0 \\
17 & S/32 & Tokens Choice & 32 & 392 & 2 & 2.0 & 48.5 & 62.8 & 9.1 & 12.1 \\
18 & S/16 & Soft MoE & 16 & 196 & -- & -- & 50.9 & 68.1 & 12.4 & 14.2 \\
19 & S/16 & Soft MoE & 32 & 196 & -- & -- & 52.0 & 70.8 & 12.9 & 14.8 \\
20 & S/16 & Dense & -- & -- & -- & -- & 47.9 & 60.8 & 17.0 & 15.3 \\
21 & S/16 & Soft MoE & 64 & 196 & -- & -- & 52.9 & 72.3 & 13.9 & 15.7 \\
22 & S/16 & Soft MoE & 128 & 196 & -- & -- & 53.6 & 73.0 & 15.9 & 17.6 \\
23 & B/32 & Experts Choice & 32 & 392 & -- & 0.5 & 49.0 & 64.3 & 13.7 & 18.0 \\
24 & B/32 & Experts Choice & 32 & 49 & -- & 0.5 & 47.8 & 62.4 & 14.3 & 18.2 \\
25 & S/16 & Experts Choice & 128 & 196 & -- & 0.5 & 50.2 & 66.7 & 15.8 & 18.5 \\
26 & S/16 & Experts Choice & 32 & 196 & -- & 1.0 & 50.5 & 67.4 & 17.5 & 18.8 \\
27 & S/16 & Experts Choice & 128 & 1568 & -- & 0.5 & 51.4 & 67.7 & 16.8 & 19.7 \\
28 & B/32 & Experts Choice & 32 & 392 & -- & 1.0 & 49.9 & 66.0 & 16.5 & 19.7 \\
29 & B/32 & Experts Choice & 64 & 392 & -- & 1.0 & 49.9 & 65.5 & 16.5 & 19.8 \\
30 & B/32 & Tokens Choice & 32 & 392 & 1 & 1.0 & 49.7 & 66.2 & 16.5 & 20.0 \\
31 & B/32 & Tokens Choice & 64 & 392 & 1 & 1.0 & 49.8 & 65.6 & 16.5 & 20.2 \\
32 & B/32 & Soft MoE & 64 & 49 & -- & -- & 51.8 & 70.7 & 17.8 & 20.3 \\
33 & B/32 & Experts Choice & 64 & 49 & -- & 1.0 & 48.6 & 64.0 & 17.7 & 20.3 \\
34 & B/32 & Experts Choice & 32 & 49 & -- & 1.0 & 48.4 & 63.8 & 17.7 & 20.5 \\
35 & S/16 & Experts Choice & 32 & 1568 & -- & 1.0 & 51.3 & 68.7 & 21.5 & 21.5 \\
36 & S/16 & Soft MoE & 256 & 196 & -- & -- & 53.8 & 73.7 & 19.9 & 22.1 \\
37 & S/16 & Tokens Choice & 32 & 1568 & 1 & 1.0 & 51.2 & 68.9 & 21.5 & 23.2 \\
38 & S/16 & Experts Choice & 256 & 196 & -- & 1.0 & 50.7 & 67.7 & 19.8 & 23.3 \\
39 & S/16 & Experts Choice & 32 & 196 & -- & 2.0 & 51.0 & 68.3 & 23.1 & 23.5 \\
40 & B/32 & Tokens Choice & 32 & 392 & 2 & 1.0 & 50.2 & 67.4 & 22.0 & 23.6 \\
41 & B/32 & Tokens Choice & 64 & 392 & 2 & 1.0 & 50.8 & 68.0 & 22.1 & 23.8 \\
42 & S/16 & Tokens Choice & 64 & 1568 & 1 & 1.0 & 51.5 & 69.1 & 21.3 & 24.9 \\
43 & S/16 & Experts Choice & 256 & 1568 & -- & 1.0 & 52.3 & 69.7 & 21.7 & 25.5 \\
44 & S/16 & Experts Choice & 32 & 1568 & -- & 2.0 & 52.4 & 70.3 & 31.0 & 27.8 \\
45 & S/16 & Tokens Choice & 32 & 1568 & 2 & 1.0 & 52.1 & 70.3 & 31.0 & 30.0 \\
46 & B/32 & Tokens Choice & 64 & 392 & 2 & 2.0 & 51.2 & 70.0 & 33.2 & 30.4 \\
47 & B/32 & Tokens Choice & 32 & 392 & 2 & 2.0 & 51.0 & 69.5 & 33.6 & 31.1 \\
48 & S/16 & Tokens Choice & 64 & 1568 & 2 & 1.0 & 52.3 & 70.4 & 31.1 & 32.0 \\
49 & S/16 & Tokens Choice & 32 & 1568 & 2 & 2.0 & 52.8 & 71.4 & 50.0 & 42.5 \\
50 & S/16 & Tokens Choice & 64 & 1568 & 2 & 2.0 & 52.9 & 71.4 & 50.1 & 45.1 \\
51 & B/16 & Dense & -- & -- & -- & -- & 52.0 & 71.8 & 64.8 & 45.2 \\
52 & B/16 & Soft MoE & 128 & 196 & -- & -- & 55.3 & 77.0 & 59.0 & 46.8 \\
53 & B/16 & Experts Choice & 128 & 1568 & -- & 0.5 & 53.7 & 73.0 & 59.0 & 48.2 \\
54 & B/16 & Experts Choice & 32 & 196 & -- & 1.0 & 53.3 & 73.0 & 65.6 & 51.0 \\
55 & B/16 & Experts Choice & 128 & 196 & -- & 0.5 & 52.5 & 72.2 & 58.8 & 52.6 \\
56 & L/32 & Dense & -- & -- & -- & -- & 51.3 & 70.9 & 55.9 & 54.9 \\
57 & L/32 & Experts Choice & 32 & 392 & -- & 0.5 & 52.3 & 71.2 & 47.4 & 55.2 \\
58 & L/32 & Experts Choice & 32 & 49 & -- & 0.5 & 51.1 & 70.6 & 49.8 & 55.7 \\
59 & L/32 & Soft MoE & 32 & 49 & -- & -- & 53.5 & 75.0 & 49.8 & 56.0 \\
60 & B/16 & Experts Choice & 32 & 1568 & -- & 1.0 & 54.2 & 74.5 & 73.6 & 56.2 \\
61 & B/16 & Tokens Choice & 32 & 1568 & 1 & 1.0 & 53.7 & 74.4 & 73.6 & 57.8 \\
62 & B/16 & Experts Choice & 256 & 196 & -- & 1.0 & 52.7 & 72.7 & 73.4 & 58.1 \\
63 & B/16 & Soft MoE & 256 & 196 & -- & -- & 55.8 & 78.0 & 73.7 & 58.2 \\
64 & B/16 & Tokens Choice & 64 & 1568 & 1 & 1.0 & 54.0 & 74.8 & 73.2 & 58.7 \\
65 & L/32 & Experts Choice & 64 & 392 & -- & 1.0 & 52.7 & 72.1 & 56.9 & 60.4 \\
66 & B/16 & Experts Choice & 256 & 1568 & -- & 1.0 & 53.9 & 73.5 & 73.8 & 60.5 \\
67 & L/32 & Experts Choice & 32 & 392 & -- & 1.0 & 52.7 & 71.7 & 56.8 & 60.6 \\
68 & L/32 & Tokens Choice & 64 & 392 & 1 & 1.0 & 51.9 & 71.4 & 56.9 & 61.0 \\
69 & L/32 & Tokens Choice & 32 & 392 & 1 & 1.0 & 52.3 & 71.7 & 57.1 & 61.6 \\
70 & L/32 & Experts Choice & 64 & 49 & -- & 1.0 & 51.1 & 70.7 & 61.6 & 62.6 \\
71 & L/32 & Soft MoE & 64 & 49 & -- & -- & 54.0 & 75.2 & 61.7 & 62.8 \\
72 & L/32 & Experts Choice & 32 & 49 & -- & 1.0 & 51.4 & 70.3 & 61.5 & 63.2 \\
73 & B/16 & Experts Choice & 32 & 196 & -- & 2.0 & 53.1 & 73.9 & 86.8 & 64.2 \\
74 & L/32 & Tokens Choice & 32 & 392 & 2 & 1.0 & 51.5 & 70.7 & 76.0 & 72.2 \\
75 & B/16 & Experts Choice & 32 & 1568 & -- & 2.0 & 54.6 & 75.6 & 102.9 & 72.5 \\
76 & L/32 & Tokens Choice & 64 & 392 & 2 & 1.0 & 52.0 & 71.8 & 76.0 & 72.5 \\
77 & B/16 & Tokens Choice & 32 & 1568 & 2 & 1.0 & 53.9 & 74.7 & 102.9 & 74.7 \\
78 & B/16 & Soft MoE & 512 & 196 & -- & -- & 56.1 & 78.5 & 103.1 & 76.5 \\
79 & B/16 & Tokens Choice & 64 & 1568 & 2 & 1.0 & 54.3 & 74.8 & 103.0 & 76.5 \\
80 & S/8 & Dense & -- & -- & -- & -- & 49.9 & 66.7 & 82.7 & 77.7 \\
81 & S/8 & Soft MoE & 512 & 784 & -- & -- & 56.1 & 78.0 & 85.6 & 88.5 \\
82 & S/8 & Experts Choice & 32 & 784 & -- & 1.0 & 52.9 & 72.6 & 91.3 & 93.0 \\
83 & L/32 & Tokens Choice & 64 & 392 & 2 & 2.0 & 52.9 & 72.9 & 114.3 & 93.2 \\
84 & L/32 & Tokens Choice & 32 & 392 & 2 & 2.0 & 52.5 & 72.5 & 115.7 & 95.8 \\
85 & L/16 & Dense & -- & -- & -- & -- & 54.8 & 77.8 & 226.9 & 100.9 \\
86 & L/16 & Experts Choice & 128 & 196 & -- & 0.5 & 54.0 & 76.7 & 204.6 & 104.9 \\
87 & L/16 & Soft MoE & 128 & 196 & -- & -- & 57.2 & 80.3 & 205.0 & 106.0 \\
88 & B/16 & Tokens Choice & 32 & 1568 & 2 & 2.0 & 54.4 & 75.7 & 161.4 & 108.4 \\
89 & B/16 & Tokens Choice & 64 & 1568 & 2 & 2.0 & 54.8 & 76.0 & 161.5 & 110.5 \\
90 & L/16 & Experts Choice & 32 & 196 & -- & 1.0 & 55.1 & 77.5 & 228.6 & 113.6 \\
91 & L/16 & Tokens Choice & 32 & 1568 & 1 & 1.0 & 55.9 & 78.5 & 250.4 & 125.1 \\
92 & L/16 & Tokens Choice & 64 & 1568 & 1 & 1.0 & 56.2 & 78.6 & 248.8 & 125.7 \\
93 & S/8 & Experts Choice & 32 & 6272 & -- & 1.0 & 53.6 & 73.4 & 160.6 & 126.6 \\
94 & S/8 & Experts Choice & 512 & 784 & -- & 1.0 & 53.4 & 72.4 & 104.1 & 129.0 \\
95 & L/16 & Soft MoE & 256 & 196 & -- & -- & 57.4 & 80.2 & 256.0 & 129.6 \\
96 & S/8 & Tokens Choice & 32 & 6272 & 1 & 1.0 & 53.8 & 73.7 & 162.5 & 129.8 \\
97 & L/16 & Experts Choice & 256 & 196 & -- & 1.0 & 54.1 & 76.7 & 255.2 & 130.1 \\
98 & L/16 & Experts Choice & 32 & 196 & -- & 2.0 & 55.2 & 77.8 & 301.0 & 140.3 \\
99 & S/8 & Experts Choice & 512 & 6272 & -- & 1.0 & 54.8 & 74.6 & 149.3 & 161.9 \\
100 & S/8 & Tokens Choice & 32 & 6272 & 2 & 1.0 & 54.2 & 74.6 & 243.4 & 166.6 \\
101 & H/14 & Soft MoE & 128 & 256 & -- & -- & 58.0 & 81.6 & 599.2 & 170.5 \\
102 & H/14 & Dense & -- & -- & -- & -- & 56.5 & 80.1 & 680.5 & 196.2 \\
103 & H/14 & Experts Choice & 64 & 2048 & -- & 1.25 & 57.3 & 80.4 & 855.9 & 210.9 \\
104 & L/16 & Tokens Choice & 32 & 1568 & 2 & 2.0 & 53.5 & 74.6 & 534.5 & 218.5 \\
105 & L/16 & Tokens Choice & 64 & 1568 & 2 & 2.0 & 53.3 & 73.3 & 535.1 & 226.9 \\
106 & H/14 & Tokens Choice & 64 & 2048 & 1 & 1.25 & 56.7 & 79.8 & 857.0 & 230.7 \\
107 & S/8 & Tokens Choice & 32 & 6272 & 2 & 2.0 & 54.1 & 74.8 & 424.4 & 255.4 \\
\end{longtable}

\end{document}